\pdfoutput=1
\documentclass[11pt]{article}

\usepackage[final]{acl}

\usepackage{makecell}
\usepackage{placeins}
\usepackage{CJKutf8}
\usepackage{times}
\usepackage{latexsym}
\usepackage{epsfig}
\usepackage{graphicx}
\usepackage{xcolor}
\usepackage{amsmath}
\usepackage{amssymb}
\usepackage{algpseudocode}

\usepackage{multirow}
\usepackage{booktabs}
\usepackage{algorithm}
\usepackage{subfigure}
\usepackage{bbm}
\usepackage{amssymb}
\usepackage{colortbl}
\usepackage{pifont}
\definecolor{mypink}{rgb}{.99,.91,.95}
\definecolor{mygreen}{rgb}{.9,.99,.9}
\definecolor{mygray}{gray}{.9}

\usepackage{tcolorbox}
\usepackage{listings}
\usepackage{enumitem}
\setenumerate[1]{itemsep=-1pt,partopsep=0pt,topsep=0pt}
\setitemize[1]{itemsep=-1pt,partopsep=0pt,topsep=0pt}
\setdescription{itemsep=-1pt,partopsep=0pt,topsep=0pt}

\usepackage[T1]{fontenc}
\usepackage[utf8]{inputenc}

\usepackage{microtype}

\usepackage{inconsolata}

\usepackage{graphicx}

\title{VidCapBench: A Comprehensive Benchmark of Video Captioning for Controllable Text-to-Video Generation}

\author{
 \textbf{Xinlong Chen\textsuperscript{1,2}}\thanks{Work done during an internship at Kuaishou Technology.},
 \textbf{Yuanxing Zhang\textsuperscript{3}},
 \textbf{Chongling Rao\textsuperscript{3}},
 \textbf{Yushuo Guan\textsuperscript{3}},
 \textbf{Jiaheng Liu\textsuperscript{4}}, \\
 \textbf{Fuzheng Zhang\textsuperscript{3}},
 \textbf{Chengru Song\textsuperscript{3}},
 \textbf{Qiang Liu\textsuperscript{1,2}}\thanks{Corresponding author: \href{qiang.liu@nlpr.ia.ac.cn}{qiang.liu@nlpr.ia.ac.cn}},
 \textbf{Di Zhang\textsuperscript{3}},
 \textbf{Tieniu Tan\textsuperscript{1,2,4}}
\\
 \textsuperscript{1}New Laboratory of Pattern Recognition (NLPR),\\
Institute of Automation, Chinese Academy of Sciences (CASIA)\\
 \textsuperscript{2}School of Artificial Intelligence, University of Chinese Academy of Sciences\\
 \textsuperscript{3}Kuaishou Technology
 \textsuperscript{4}Nanjing University
\\
}

\begin{document}
\maketitle
\begin{abstract}
The training of controllable text-to-video (T2V) models relies heavily on the alignment between videos and captions, yet little existing research connects video caption evaluation with T2V generation assessment.
This paper introduces VidCapBench, a video caption evaluation scheme specifically designed for T2V generation, agnostic to any particular caption format.
% VidCapBench curates a dataset of 643 video clips selected across multiple dimensions, including subject type, subject count, artistic style, and motion intensity.
VidCapBench employs a data annotation pipeline, combining expert model labeling and human refinement, to associate each collected video with key information spanning video aesthetics, content, motion, and physical laws.
VidCapBench then partitions these key information attributes into automatically assessable and manually assessable subsets, catering to both the rapid evaluation needs of agile development and the accuracy requirements of thorough validation.
By evaluating numerous state-of-the-art captioning models, we demonstrate the superior stability and comprehensiveness of VidCapBench compared to existing video captioning evaluation approaches.
Verification with off-the-shelf T2V models reveals a significant positive correlation between scores on VidCapBench and the T2V quality evaluation metrics, indicating that VidCapBench can provide valuable guidance for training T2V models. The project is available at \url{https://github.com/VidCapBench/VidCapBench}.
\end{abstract}

\section{Introduction}

% Controllable text-to-video (T2V) generation has recently witnessed significant improvement~\cite{singer2022make,wu2023tune,zhang2023controlvideo}. 
% This technology leverages text prompts to control video synthesis~\cite{genmo2024mochi,kong2024hunyuanvideo,zhou2024allegro}, enabling the instant visualization of designs and facilitating applications in creative content production and entertainment.
Controllable text-to-video (T2V) generation leverages text prompts to guide video synthesis~\cite{genmo2024mochi,zhou2024allegro}, enabling the instant visualization of designs and facilitating applications in creative content and entertainment.
Advances in generative model's backbones~\cite{blattmann2023stable,esser2024scaling,peebles2023scalable,weng2024art} further innovate the video generation process to adhere to textual instructions, exhibit aesthetic appeal, and conform to physical laws.
Video captioning, the crucial supporting infrastructure to T2V generation, has also progressed.
% Increasingly, researchers recognize that video captions are primarily for machine consumption instead of human interpretation.
Coarse-grained or detail-lacking captions significantly hinder both the comprehension and reconstruction of visual information~\cite{jin2024pyramidal,cheng2024videgothink}. 
Hence, prevalent T2V models are devoted to strengthening the alignment between the generated content and the detailed prompts/captions~\cite{kim2023dense,liu2024improving}.
With the objective to optimize this alignment, T2V models present high fidelity in subjects' motion~\cite{wei2024dreamvideo,wang2024dreamrunner,zhou2024motion}, temporal changes~\cite{guo2024trace,yang2023vid2seq,xiong2024lvd}, and event progression~\cite{he2024storyteller,wang2024tarsier}.
% Furthermore, captions facilitate the construction of multi-faceted evaluation metrics for assessing the semantic alignment and objective reasonableness of generated videos.
Meanwhile, the quality of datasets used for training captioning models~\cite{hong2024cogvlm2,zhang2023video} has also improved considerably, such as OpenVid~\cite{nan2024openvid} and ShareGPT4V~\cite{chen2025sharegpt4v}.
% DOCCI~\cite{onoe2025docci}, DenseFusion~\cite{li2024densefusion}, BLIP3-Kale~\cite{awadalla2024blip3}, OpenVid~\cite{nan2024openvid}, ShareGPT4V~\cite{chen2025sharegpt4v}, and Koala~\cite{wang2024koala}.

\begin{figure}
  \includegraphics[width=\linewidth, height=5.74cm]{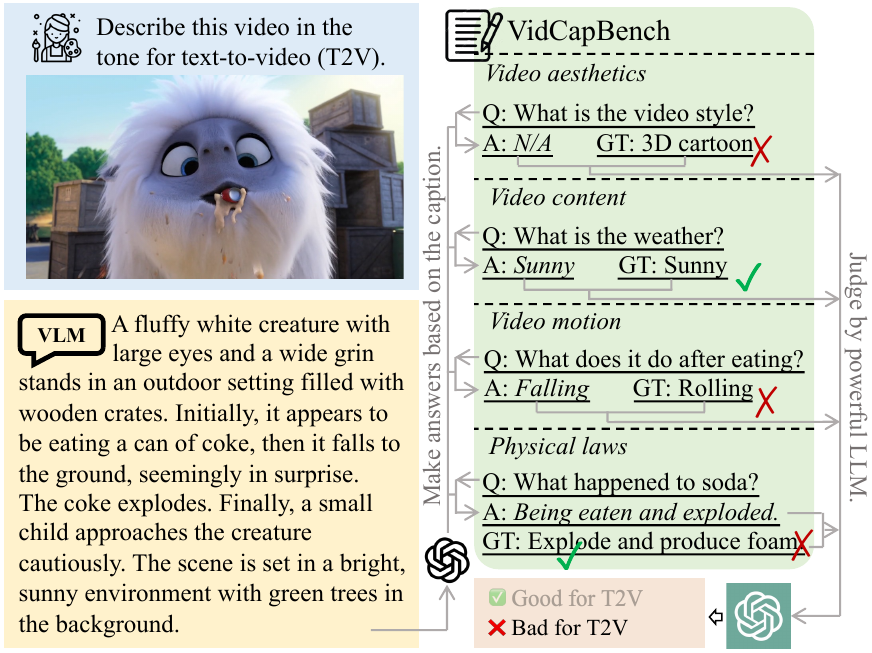}
  \caption{VidCapBench evaluates the video captioning model from the aspects of T2V generation.}
  \label{fig:teaser}
\end{figure}

\begin{table*}
    \centering
    \resizebox{\linewidth}{!}{
    \begin{tabular}{l c c c c c c c c c c}
    \toprule
        Benchmark & Metrics & \# Videos & \# QA pairs & Video diversity & Aesthetics & Subject & Motion & Physical law & Conciseness & Caption format \\
        \midrule
        MSR-VTT~\cite{xu2016msr} & CIDEr & 2,990 & 2,990 & \ding{55} & \ding{55} & \ding{55} & \ding{55} & \ding{55} & \ding{52} & Short \\
        VATEX~\cite{wang2019vatex} & CIDEr & 4,478 & 4,478 & \ding{55} & \ding{55} & \ding{55} & \ding{55} & \ding{55} & \ding{52} & Short \\
        DREAM-1K~\cite{wang2024tarsier} & Pre/Rec/F1 & 1,000 & 6,298 & \ding{52} & \ding{55} & \ding{52} & \ding{55} & \ding{55} & \ding{55} & Unstructured \\
        VDC~\cite{chai2024auroracap} & Acc/VDCScore & 1,027 & 96,902 & \ding{55} & \ding{55} & \ding{52} & \ding{52} & \ding{55} & \ding{55} & Structured \\
        \midrule
        VidCapBench & Acc/Pre/Cov/Con & 643 & 10,644 & \ding{52} & \ding{52} & \ding{52} & \ding{52} & \ding{52} & \ding{52} & Arbitrary \\
        \bottomrule
    \end{tabular}
    }
    \caption{Comparison between VidCapBench and mainstream video caption benchmarks.}
    \label{tab:teasor_comparison}
\end{table*}

% Evaluating video captions significantly guides the optimization of text-to-video (T2V) models.
% As T2V models achieve a certain level of fidelity, visual realism and aesthetics become less differentiating.
% The focus shifts towards semantic responsiveness, such as the model's ability to generate specific characters and actions based on the provided text.
% Current T2V models primarily emphasize temporality, multi-agent interaction, camera movement, and style.
% These capabilities depend on the alignment between videos and captions within the training data.
To direct the optimization of T2V models, video captions should be accurate, comprehensive, diverse, concise, and abundant.
While caption formats vary across different models~\cite{ju2024miradata,zheng2024videogen}, the core elements in captions emphasized by T2V models seem to be converging.
% Existing open-source video captioning evaluation~\cite{yang2024vript,chai2024auroracap,xu2016msr} methods often prioritize summarization over video generation, providing limited guidance for the T2V training pipeline. 
A practical evaluation of video captions for T2V generation must address three main challenges:

\begin{itemize}[leftmargin=*]
\item \textbf{Alignment with T2V evaluation}: The evaluation should assess whether a video caption adequately covers aesthetics, content, motion, and physical laws, aligning with the key metrics of T2V generation.
\item \textbf{Diversity and stability}: The diversity of evaluation data and the stability of the evaluation approach influence the accurate assessment of caption quality.
\item \textbf{Impact on T2V generation}: The correlation between caption evaluation and T2V performance remains unexplored, lacking evidence on how captions influence the generated videos.
\end{itemize}

To address these challenges, we introduce VidCapBench, the first evaluation benchmark designed for assessing video captions in controllable T2V generation, as depicted in Figure~\ref{fig:teaser}.
Comparison between VidCapBench and several publicly available video caption benchmarks is presented in Table~\ref{tab:teasor_comparison}.
% Table~\ref{tab:teasor_comparison} summarizes the differences between VidCapBench and several publicly available video caption benchmarks.
Prioritizing the video diversity, VidCapBench comprises 643 richly annotated video clips.
These videos are annotated with critical aspects relevant to T2V generation, and we construct a discerning set of question-answer pairs decoupled from specific caption formats.
The workflow of VidCapBench is transferable to arbitrary in-house datasets for the more targeted evaluation. The main contributions of this paper are summarized as follows:

\begin{itemize}[leftmargin=*]
\item We introduce VidCapBench, a novel benchmark designed to facilitate comprehensive and stable evaluation of video captions across multiple dimensions relevant to T2V generation.
\item We propose a two-stage evaluation method: rapid automated evaluation on a stable-to-judge subset provides quick feedback for developers, while introducing accurate human evaluation on the remaining subset offers crucial guidance.
\item Our experiments demonstrate that most open-source captioning models perform inferior to proprietary models like GPT-4o.
Applying captions to several production-ready T2V models reveals a strong positive correlation between the performance on VidCapBench and the quality of generated videos, validating the effectiveness of our proposed evaluation approach.
\end{itemize}

\section{Related Work}

\noindent\textbf{Video captioning.}
The goal of video captioning is to describe a video across several key aspects, aiding understanding~\cite{doveh2023dense}, retrieval~\cite{ma2024drvideo}, and motion control~\cite{wang2024motionctrl}.
In T2V generation, accurate and detailed video captions can enhance semantic alignment during model training~\cite{polyak2024movie}.
Naive captioning models adopt free-form descriptions~\cite{chen2024panda,wang2024koala}.
To enhance controllability, MiraData~\cite{ju2024miradata}, VDC~\cite{chai2024auroracap}, and Vript~\cite{yang2024vript} emphasize specific aspects like subjects, background, and shots, significantly benefiting T2V generation.
Other methods describe videos from an event perspective~\cite{wang2024tarsier,he2024storyteller}, capturing temporal information more effectively.
Despite advancements in caption controlling~\cite{wang2023caption,hua2024finecaption}, 
% metrics on evaluating the caption model often remain limited.
evaluations with omissions may lead to a seesaw effect where gains in one dimension come at the cost of others, limiting the utility of the captioning model.

\noindent\textbf{Evaluation methods for video captioning.}
% While the effective evaluation of video captions was previously underemphasized, the advancement of text-to-video generation has spurred the development of increasingly sophisticated evaluation metrics, constantly evolving alongside changes in caption formats.
The advancement of T2V generation has spurred the development of evaluation approaches for video captioning.
Traditional approaches~\cite{xu2017video,xu2016msr} for short captions rely on legacy metrics like CIDEr and BLEU.
For dense captions, inspired by image captioning evaluation~\cite{liu2024playground,prabhu2024trust,tu2024automatic}, many approaches employ question answering (QA) followed by natural language inference (NLI) with a critic model.
Existing evaluation schemes of video captions are often tied to specific caption formats and suffer from instability in automatic evaluation.
In this context, VidCapBench emerges as a more robust solution, offering a comprehensive and stable evaluation framework that aligns with the controllable T2V evaluation~\cite{rawte2024vibe,huang2024vbench++,he2024videoscore}, providing better guidance for T2V model training.

\section{VidCapBench}

In this section, we introduce the design and curation of VidCapBench.

\subsection{Preliminaries}

Caption evaluation is typically performed through human or machine evaluation.

\noindent\textbf{Human evaluation}.
Human evaluation demands annotators to assess captions based on predefined criteria.
Experienced annotators deliver accurate and consistent evaluations, along with analysis of erroneous cases, which helps training T2V models.
Currently, human annotation primarily employs two methods:
% In \textit{5-point Likert scale}, annotators rate captions on a 5-point scale (1 being worst, 5 being best) based on ground truth.
% Regarding \textit{candidate comparison}, pairwise samples are presented to annotators to determine the ratio of good-same-bad and finally conclude with the ELO ratings.

\begin{itemize}[leftmargin=*]
\item \textit{5-point Likert scale}: Annotators rate captions on a 5-point scale (1: worst, 5: best) based on ground truth. Each evaluation dimension is assessed independently, with predefined examples illustrating different score levels. To ensure reliability, each example is typically evaluated by a minimum of three annotators, with inter-annotator agreement metrics employed to maintain consistency.
\item \textit{Pairwise comparison}: Annotators compare two anonymized model outputs for each example, selecting ``Caption A is better'', ``Caption B is better'', or ``Equal quality''. Pairwise comparisons typically utilize the good-same-bad metric, while ranking multiple models can employ ELO scores.
\end{itemize}

\noindent\textbf{Machine evaluation}.
Human evaluation can be inconsistent among inexperienced annotators and is generally slower and more expensive.
Conversely, automatic machine evaluation is faster and can provide some guidance for training.
Mainstream machine evaluation often utilizes GPT-4 as a judge, which can be divided into two categories:
\begin{itemize}[leftmargin=*]
\item \textit{Predefined-QA paradigm}: Multiple key information points are annotated for each video by QA pairs.
Captions are evaluated by posing questions to the judge model, awarding points only for correct answers.
Natural Language Inference (NLI) is used to categorize answers as ``Entailed'', ``Neutral'', or ``Contradictory''.
\item \textit{Retrieval-based paradigm}: This approach generates a series of yes/no questions about entities based on a given caption~\cite{cho2023davidsonian}.
A judge model then answers these questions using the original video as context.
Descriptions corresponding to questions answered with ``no'' are considered hallucinatory.
Note that this approach may incur high computational costs due to the repeated video question-answering process.
\end{itemize}

\subsection{Benchmarking Video Captions}~\label{sec:alignment}
% To provide a robust evaluation scheme for video captions, VidCapBench addresses two fundamental questions: what to evaluate and how to evaluate.
To establish a comprehensive evaluation framework for video captions, VidCapBench tackles two fundamental inquiries: what criteria should be employed to align the caption evaluation with T2V generation, and how to ensure a stable and reliable evaluation process.

\noindent\textbf{Alignment with T2V evaluation}.
An effective T2V model is expected to produce videos with high visual fidelity, coherent object representation, precise semantic alignment with the input textual description, and realistic detail enhancement.
Correspondingly, VidCapBench evaluates video captions across the following dimensions:
\begin{itemize}[leftmargin=*]
\item \textit{Video aesthetics} (VA) encompasses the artistic and technical aspects of video creation, from filming techniques to post-production.
\item \textit{Video content} (VC) refers to the narrative content presented in the video.
\item \textit{Video motion} (VM) covers movements of foreground subjects and background objects.
\item \textit{Physical laws} (PL) allow for more realistic or dramatic visual expression, even though creativity can somewhat bend them.
\end{itemize}
Each dimension is further subdivided into specific sub-categories to ensure comprehensive and systematic evaluation coverage. The detailed categorization is provided in Appendix~\ref{sec:detail_dimensions}.

\begin{figure*}[ht]
  \includegraphics[width=\linewidth]{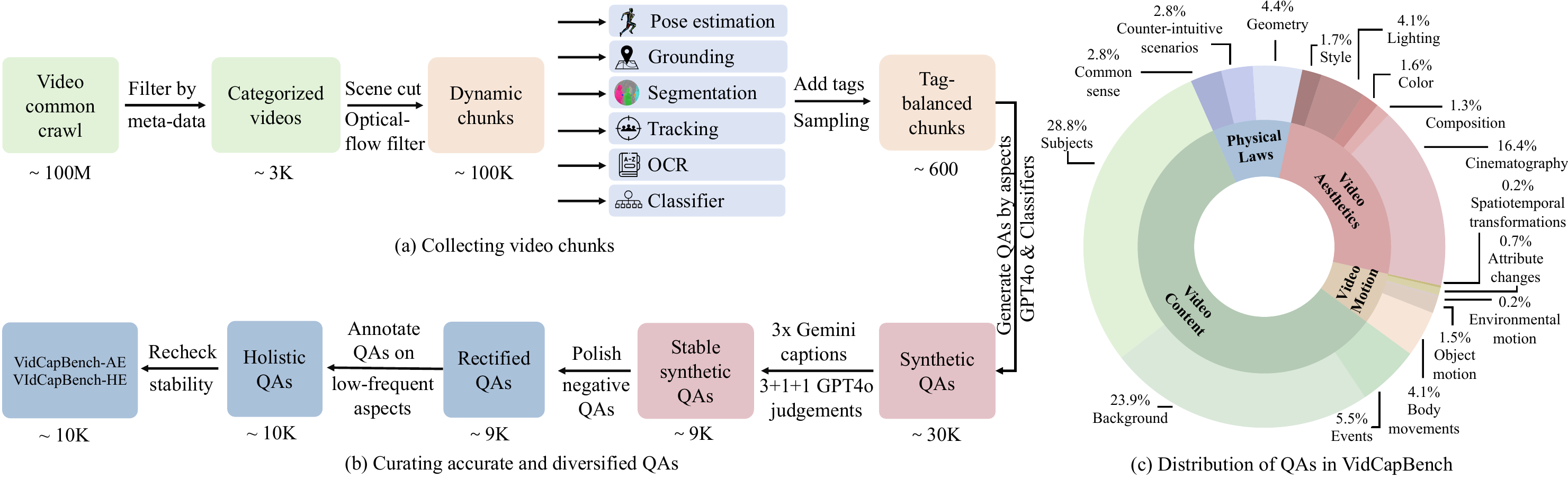}
  \caption{Illustration of the data curation pipeline and the distribution of QA pairs in VidCapBench. The QA pairs are carefully rectified to ensure that they primarily assess the quality of video captions rather than the inherent capabilities of the judge model.}
  \label{fig:pipeline}
\end{figure*}

\begin{table}
    \centering
    \resizebox{\columnwidth}{!}{
    \begin{tabular}{l | c c c c c c c}
    \toprule
        \textbf{Model} & \textbf{Eval. set} & \textbf{Overall} & \textbf{Detailed} & \textbf{Camera} & \textbf{Short} & \textbf{Background} & \textbf{Object} \\
        \midrule
        \multirow{2}{*}{GPT-4o} & Full & \phantom{**}46.1\phantom{$_{(\textbf{+0.3})}$} & \phantom{**}50.1\phantom{$_{(\textbf{+0.3})}$} & \phantom{**}52.5\phantom{$_{(\textbf{+0.3})}$} & \phantom{***}34.3\phantom{$_{(\textbf{+0.3})}$} & 44.6\phantom{$_{(\textbf{+0.3})}$} & \phantom{**}48.8\phantom{$_{(\textbf{+0.3})}$} \\
        ~ & Selected & \phantom{**}46.4$_{(\textbf{+0.3})}$ & \phantom{**}53.1$_{(\textbf{+3.0})}$ & \phantom{**}55.5$_{(\textbf{+3.0})}$ & \phantom{***}26.7$_{(\textbf{-7.6})}$ & 46.7$_{(\textbf{+2.1})}$ & \phantom{**}52.8$_{(\textbf{+4.0})}$ \\
        \midrule
        \multirow{2}{*}{Gemini 1.5 Pro} & Full & \phantom{**}40.5\phantom{$_{(\textbf{+0.3})}$} & \phantom{**}46.3\phantom{$_{(\textbf{+0.3})}$} & \phantom{**}40.9\phantom{$_{(\textbf{+0.3})}$} & \phantom{***}30.0\phantom{$_{(\textbf{+0.3})}$} & 41.2\phantom{$_{(\textbf{+0.3})}$} & \phantom{**}44.3\phantom{$_{(\textbf{+0.3})}$} \\
        ~ & Selected & \phantom{**}44.6$_{(\textbf{+4.1})}$ & \phantom{**}52.5$_{(\textbf{+6.2})}$ & \phantom{**}48.4$_{(\textbf{+7.5})}$ & \phantom{***}25.3$_{(\textbf{-4.7})}$ & 46.2$_{(\textbf{+5.0})}$ & \phantom{**}52.4$_{(\textbf{+8.1})}$ \\
        \midrule
        \multirow{2}{*}{Qwen2-VL-72B} & Full & \phantom{**}40.0\phantom{$_{(\textbf{+0.3})}$} & \phantom{**}43.6\phantom{$_{(\textbf{+0.3})}$} & \phantom{**}46.9\phantom{$_{(\textbf{+0.3})}$} & \phantom{***}28.0\phantom{$_{(\textbf{+0.3})}$} & 39.9\phantom{$_{(\textbf{+0.3})}$} & \phantom{**}40.3\phantom{$_{(\textbf{+0.3})}$} \\
        ~ & Selected & \phantom{**}42.5$_{(\textbf{+2.5})}$ & \phantom{**}49.0$_{(\textbf{+5.4})}$ & \phantom{**}51.5$_{(\textbf{+4.6})}$ & \phantom{***}22.9$_{(\textbf{-5.1})}$ & 43.9$_{(\textbf{+4.0})}$ &\phantom{**}47.8$_{(\textbf{+7.5})}$ \\
        \midrule
        \multirow{2}{*}{CogVLM2-Caption} & Full & \phantom{**}42.8\phantom{$_{(\textbf{+0.3})}$} & \phantom{**}44.7\phantom{$_{(\textbf{+0.3})}$} & \phantom{**}47.7\phantom{$_{(\textbf{+0.3})}$} & \phantom{***}30.5\phantom{$_{(\textbf{+0.3})}$} & 45.3\phantom{$_{(\textbf{+0.3})}$} & \phantom{**}45.8\phantom{$_{(\textbf{+0.3})}$} \\
        ~ & Selected & \phantom{**}46.2$_{(\textbf{+3.4})}$ & \phantom{**}50.9$_{(\textbf{+6.2})}$ & \phantom{**}54.7$_{(\textbf{+7.0})}$ & \phantom{***}26.2$_{(\textbf{-4.3})}$ & 50.1$_{(\textbf{+4.8})}$ & \phantom{**}52.0$_{(\textbf{+6.2})}$ \\
        \midrule
        \multirow{2}{*}{Tarsier-34B} & Full & \phantom{**}37.4\phantom{$_{(\textbf{+0.3})}$} & \phantom{**}40.3\phantom{$_{(\textbf{+0.3})}$} & \phantom{**}43.1\phantom{$_{(\textbf{+0.3})}$} & \phantom{***}25.0\phantom{$_{(\textbf{+0.3})}$} & 39.5\phantom{$_{(\textbf{+0.3})}$} & \phantom{**}39.3\phantom{$_{(\textbf{+0.3})}$} \\ 
        ~ & Selected & \phantom{**}42.6$_{(\textbf{+5.2})}$ & \phantom{**}48.0$_{(\textbf{+7.7})}$ & \phantom{**}51.8$_{(\textbf{+8.7})}$ & \phantom{***}22.6$_{(\textbf{-2.4})}$ & 46.8$_{(\textbf{+7.3})}$ & \phantom{**}46.6$_{(\textbf{+7.3})}$ \\
        \bottomrule
    \end{tabular}
    }
    \caption{Accuracy comparison between the full set and the selected set which receive consistently stable evaluations on the VDC benchmark.}
    \label{tab:VDC_Consis_Compare}
\end{table}

\noindent\textbf{Stability of evaluation}.
Both the judge model's capabilities and the difficulty of evaluating the QA pairs influence the stability of machine evaluation.
For details on the former, please refer to Appendix~\ref{sec:discuss of judge}. Here, we focus on the latter.
Taking the VDC benchmark as an example, which contains roughly 100 QA pairs per video, we evaluate five models three times with GPT-4o under different random seeds. To analyze the evaluation stability of the QA pairs, we examine the number of times that the evaluation results are consistent across all three trials in the five models. 
Experimental results reveal that only 41\% of the questions receive consistent evaluations, while 13\% exhibit agreement at most twice out of five.
Furthermore, we compare the accuracy (Acc) of five captioning models on a subset of all-agreed QA pairs with that on the full VDC benchmark.
% As shown in Table~\ref{tab:VDC_Consis_Compare}, performance on the selected subset significantly surpasses that on the full set across the dimensions of ``detailed'', ``camera'', ``background'', and ``object''.
% Note that the lower Acc in the ``short'' dimension can be attributed to a discrepancy between the captioning models' outputs and VDC's expected format.
As shown in Table~\ref{tab:VDC_Consis_Compare}, the performance on the selected subset demonstrates significant discrepancies compared to that on the full benchmark, which highlights the unreliability of evaluating all QA pairs solely through automated methods. Instead, a more refined approach is warranted, wherein QA pairs should be categorized into two groups: (1) those suitable for automated evaluation due to their high machine evaluation consistency, and (2) the remaining, more challenging QA pairs that necessitate human intervention for nuanced differentiation.
% This observation suggests that a subset of questions with high evaluation consistencies enables reliable machine evaluation, while others may necessitate human intervention for accurate evaluation. Consequently, QA pairs can be categorized into two groups: (1) those suitable for automated evaluation due to their high machine evaluation consistency, and (2) more challenging ones requiring human evaluation for nuanced differentiation.

\noindent\textbf{Metrics of caption evaluation}.
% A good video caption should be factually accurate, clearly expressed, avoid rare linguistic constructions, and be easily understood by models.  VidCapBench considers the following metrics:
% \begin{itemize}[leftmargin=*]
% \item \textbf{Accuracy}: Faithfulness and correctness of the caption's content, ensuring it is reliable, verifiable, and free of hallucinations.
% \item \textbf{Clarity}: Completeness and coherence of the caption, ensuring it covers all relevant video content effectively.
% \item \textbf{Elegance}: Clarity and conciseness of expression, prioritizing well-organized, logically structured language, accurate word choice, and avoidance of redundancy, while considering text encoder length limitations.
% \end{itemize}
Considering efficiency, cost, and stability, VidCapBench employs a predefined-QA paradigm to evaluate video captioning models.
Due to the complexity of video content, answers may involve multiple adjectives, nouns, or verbs.
Therefore, the judge model is required to categorize responses into four classes: 
% Wrong ($n_w$), indicating explicit mention of the relevant content but with factual errors;
% Neutral ($n_n$), indicating no mention of the relevant content; 
% Partially correct ($n_p$), indicating mention of relevant and correct information, but with incomplete descriptions; 
% and Correct ($n_c$), indicating complete alignment between the caption and the ground-truth answer.
\begin{itemize}[leftmargin=*]
\item \textbf{Wrong} ($n_w$): Responses that explicit mention of the relevant content but with factual errors.
\item \textbf{Neutral} ($n_n$): Responses that omit the relevant content entirely.
\item \textbf{Partially correct} ($n_p$): Responses that include relevant and correct information, but with incomplete descriptions.
\item \textbf{Correct} ($n_c$): Responses that fully align with the ground-truth answer.
\end{itemize}

Based on these classifications, we compute the following four metrics to comprehensively assess the performance of captioning models:

\begin{itemize}[leftmargin=*]
\item \textbf{Accuracy} (Acc): The proportion of responses marked as Correct, defined as $\frac{n_c}{n_p+n_c+n_w+n_n}$, reflecting the model's ability to cover comprehensive details of the video. Notably, models generating longer captions may have advantages.
\item \textbf{Precision} (Pre): Calculated as $\frac{n_p+n_c}{n_p+n_c+n_w}$, representing the proportion of mentioned content that is at least partially correct.
\item \textbf{Coverage} (Cov): Calculated as $\frac{n_p+n_c+n_w}{n_p+n_c+n_w+n_n}$, representing the proportion of addressed content relative to the total content covered by the QA pairs of the video.
\item \textbf{Conciseness} (Con): Measured by the contribution of each text token to Acc, defined as Acc/$\tau$, where $\tau$ represents the token number of the corresponding captions, as determined by a T5 model~\cite{raffel2020exploring}.
\end{itemize}

% \begin{figure}[t]
%   \includegraphics[width=\columnwidth]{agreement acc compare/Gemini 1.5 Pro_category_acc_comparison.pdf}
%   \includegraphics[width=\columnwidth]{agreement acc compare/Tarsier-34B_category_acc_comparison.pdf}
%   \caption{Accuracy comparison between the full set and the selected set where questions received consistently stable evaluations.}
%   \label{fig:Gemini & tarsier acc compare}
% \end{figure}

\begin{figure*}[!h]
  \includegraphics[width=\linewidth, height=9.1cm]{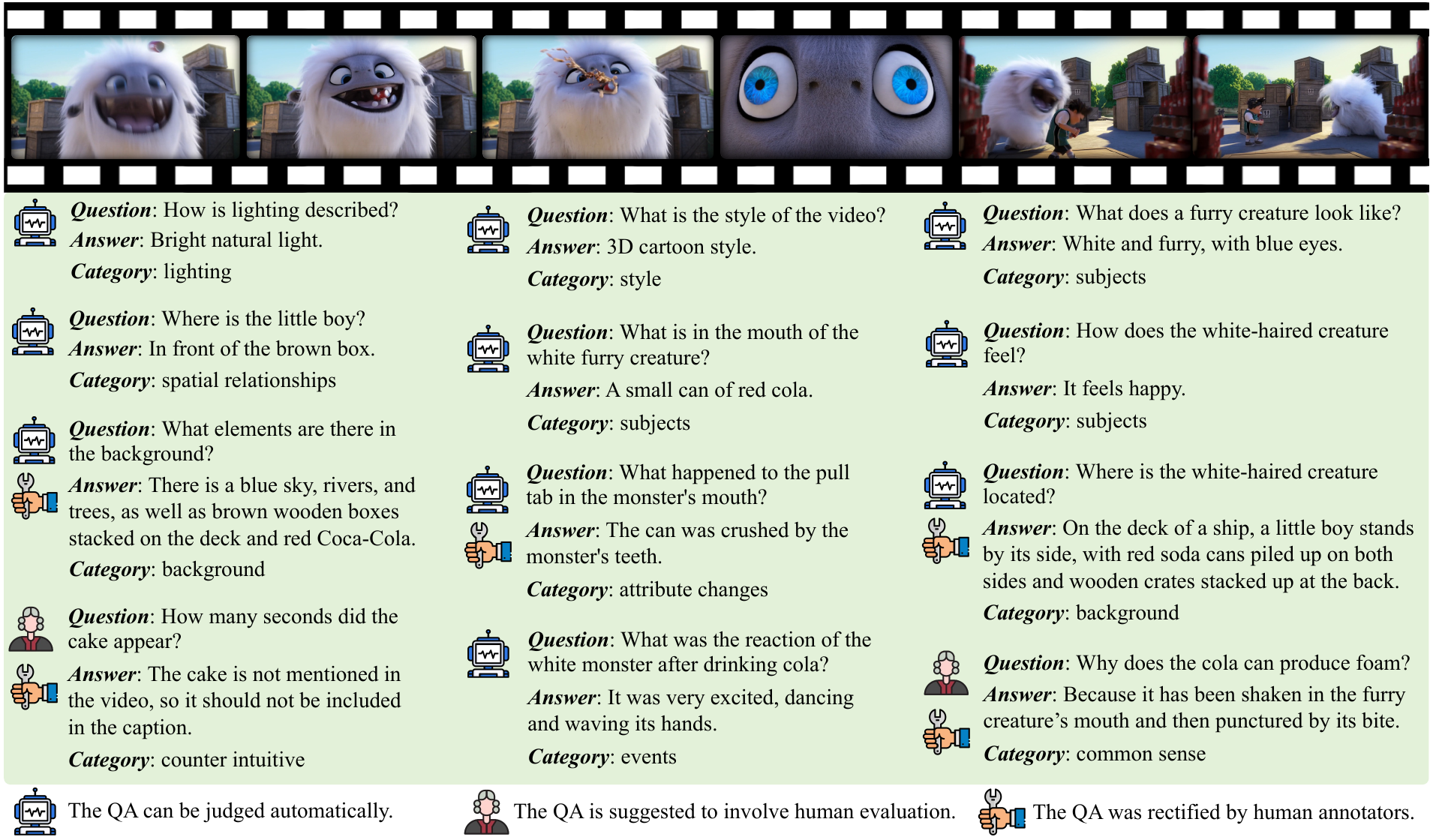}
  \caption{An example of the QA pairs for a video in VidCapBench.}
  \label{fig:demo_qa}
\end{figure*}

% \begin{figure*}[!h]
%   \includegraphics[width=\linewidth, height=9.57cm]{figures_v2/demo_qa_2.pdf}
%   \caption{An example of the QA pairs for a video in VidCapBench.}
%   \label{fig:demo_qa}
% \end{figure*}

\subsection{Data Curation}

Open-source video datasets are often delicately curated and valuable for detailed analysis. 
Hence, we sample videos from several prominent open-source datasets. 
However, recognizing that these videos may have been extensively captioned and incorporated in many training datasets, we augment our data collection with additional copyright-free videos from YouTube and public user-generated content (UGC) platforms, extracting segments to ensure a portion of our data remains unexposed to prior training or processing.
Figure~\ref{fig:pipeline} illustrates the pipeline of curating VidCapBench.

\subsubsection{Videos Collection}

We focus on a set of subjects in our analysis: person, animal, plant, food, common object, landscape, vehicle, building, specific intellectual property (IP), and no subject. 
Uniform involvement is expected across these categories.
We perform an initial filtering based on captions provided by the Omega-MultiModal project~\footnote{Huggingface: omegalabsinc/omega-multimodal}, employing Qwen2-VL-72B~\cite{wang2024qwen2} for classification, retaining approximately 3,000 videos per category.  
These videos are then segmented into 3-15 second clips using PySceneDetect~\footnote{https://github.com/Breakthrough/PySceneDetect}.  
Videos made of static images are removed using an optical flow tool.  Subsequently, we sample 16 frames from each video and perform the following operations in parallel:
Pose estimation~\cite{khirodkar2025sapiens} to detect human presence and pose variations;
Object detection and grounding~\cite{liu2025grounding} over the initial keywords to identify the amount of the target subject;
Object tracking~\cite{zhang2022bytetrack} to track object motion, labeling objects as static or consistently trackable;
Image segmentation~\cite{ravi2024sam} to filter excessively complex scenes;
Optical character recognition (OCR)~\cite{wei2024general} to identify and minimize the presence of text.
% \begin{itemize}[leftmargin=*]
% \item Pose Estimation: We used sapiens-pose-1b~\cite{khirodkar2025sapiens} to detect human presence and pose variations within the sampled frames.
% \item Object Detection and Grounding: Using keywords from the initial classification, we applied GroundingDINO-base~\cite{liu2025grounding} to the sampled frames, ensuring the target subject appeared in at least 10 frames with sufficient relative size, and recorded the number of detected objects.
% \item Object Tracking: We employed bytetrack\_x\_mot20~\cite{zhang2022bytetrack} to track object motion, labeling objects as static or consistently trackable.
% \item Image Segmentation: We used sam2.1-hiera-large~\cite{ravi2024sam} for image segmentation.  Considering practical text-to-video generation requirements, we recorded the number of segmented regions to filter excessively complex scenes (retaining 16-48 segments in practical applications).
% \item Optical Character Recognition: We used a lightweight OCR model~\cite{wei2024general} to detect text within the frames.  We aimed to minimize the presence of text, particularly subtitles, prioritizing naturally occurring characters.
% \end{itemize}
These generated labels inform the subsequent balanced sampling. Furthermore, we employ optical flow models, a custom-trained artistic style classifier, and a set of custom-trained character attribute classifiers for additional video labeling.
Finally, we randomly select videos and manually verify the presence of each label at least three times, ensuring uniform distribution across labels.

\begin{table*}
    \centering
    \resizebox{\linewidth}{!}{
    \begin{tabular}{lcccccc|c}
    \toprule
        \multirow{3}{*}{\textbf{Model}} & \multirow{3}{*}{\textbf{Frames}} & \multicolumn{5}{c|}{\textbf{VidCapBench-AE}} & \textbf{VDC} \\
        ~ & ~ & \textbf{Overall} & \textbf{Video Aesthetics} & \textbf{Video Content} & \textbf{Video Motion} & \textbf{Physical Laws} & \textbf{Overall} \\
        ~ & ~ & \textbf{Acc / Pre / Cov / Con} & \textbf{Acc / Pre / Cov / Con} & \textbf{Acc / Pre / Cov / Con} & \textbf{Acc / Pre / Cov / Con} & \textbf{Acc / Pre / Cov / Con} & \textbf{Acc / Score} \\
        \toprule
        GPT-4o-20240806 & 16 & 16.8 ($\pm$ 0.1) / 57.4 / 86.0 / ~5.9~ & 14.1 / 47.6 / 83.4 / ~4.9~ & 17.5 / 61.7 / 87.2 / ~6.1~ & 10.2 / 41.3 / 84.0 / ~3.6~ & 27.9 / 62.1 / 85.4 / ~9.7~ & 46.1 ($\pm$ 1.0) / 2.2 \\
        Gemini-1.5-Pro-002 & - & 17.1 ($\pm$ 0.2) / 54.8 / 87.4 / ~9.2~ & 16.4 / 47.6 / 85.4 / ~8.8~ & 16.9 / 57.8 / 88.5 / ~9.1~ & ~9.8~ / 45.1 / 80.9 / ~5.3~ & 28.4 / 59.3 / 88.2 / 15.3 & 40.5 ($\pm$ 1.6) / 2.0 \\
        \midrule
        Llava-Next-Video-7B~\cite{zhang2024llava} & 16 & 10.6 ($\pm$ 0.1) / 42.3 / 79.4 / ~3.9~ & 11.3 / 39.9 / 82.2 / ~4.2~ & ~9.6~ / 43.2 / 78.1 / ~3.5~ & ~4.4~ / 23.7 / 75.1 / ~1.7~ & 24.4 / 54.5 / 82.9 / ~9.0~ & 37.6 ($\pm$ 2.2) / 1.9 \\
        LongVA-7B~\cite{zhang2024longcontexttransferlanguage} & 128 & 10.8 ($\pm$ 0.1) / 43.0 / 79.3 / ~6.1~ & 12.8 / 42.1 / 83.8 / ~7.3~ & ~9.2~ / 43.4 / 77.2 / ~5.2~ & ~4.9~ / 25.1 / 79.6 / ~2.8~ & 24.9 / 52.9 / 83.2 / 14.1 & 36.1 ($\pm$ 1.9) / 1.9 \\
        mPLUG-Owl3-7B~\cite{ye2024mplug} & 16 & 14.5 ($\pm$ 0.3) / 49.6 / 84.4 / ~6.9~ & 12.9 / 40.7 / 83.7 / ~6.1~ & 14.8 / 53.5 / 85.1 / ~7.0~ & ~5.3~ / 33.3 / 80.0 / ~2.5~ & 26.9 / 55.7 / 81.7 / 12.8 & 36.5 ($\pm$ 2.7) / 1.9 \\
        InternVL2-8B~\cite{chen2024far} & 32 & 10.2 ($\pm$ 0.2) / 43.0 / 84.9 / ~2.5~ & ~9.1~ / 36.3 / 84.4 / ~2.2~ & 10.0 / 46.1 / 85.2 / ~2.4~ & ~4.4~ / 18.0 / 81.3 / ~1.1~ & 23.6 / 52.8 / 85.7 / ~5.8~ & 37.4 ($\pm$ 2.4) / 1.9 \\
        Qwen2-VL-7B~\cite{wang2024qwen2} & 2 fps & 11.1 ($\pm$ 0.2) / 47.1 / 77.0 / ~6.4~ & 12.4 / 44.3 / 78.7 / ~7.2~ & ~9.9~ / 48.3 / 75.9 / ~5.7~ & ~4.0~ / 22.7 / 78.2 / ~2.3~ & 26.1 / 59.4 / 81.2 / 15.1 & 39.6 ($\pm$ 1.6) / 2.0 \\
        % Llava-Video-7B~\cite{zhang2024video} & 32 & 12.6 ($\pm$ 0.6) / 35.4 / 87.9 / ~3.1~ & ~7.5~ / 26.9 / 75.2 / ~9.4~ & 14.4 / 39.3 / 93.7 / 12.9 & ~5.4~ / 27.2 / 84.7 / ~8.5~ & 17.1 / 42.2 / 87.9 / 28.1 \textbf{TODO} \\
        \midrule
        Pixtral-12B~\cite{agrawal2024pixtral} & 16 & 11.0 ($\pm$ 0.3) / 39.5 / 79.6 / ~5.2~ & 14.5 / 42.7 / 82.8 / ~6.8~ & ~8.6~ / 37.9 / 78.4 / ~4.0~ & ~3.6~ / 18.6 / 69.3 / ~1.7~ & 28.6 / 52.4 / 82.4 / 13.5 & 39.0 ($\pm$ 2.3) / 2.0 \\
        CogVLM2-Caption~\cite{hong2024cogvlm2} & 1 fps & 13.1 ($\pm$ 0.2) / 49.2 / 85.1 / ~8.4~ & 12.5 / 45.2 / 83.1 / ~8.0~ & 12.7 / 50.8 / 86.3 / ~8.1~ & ~5.7~ / 33.9 / 82.7 / ~3.7~ & 27.9 / 59.9 / 82.7 / 17.8 & 42.8 ($\pm$ 1.4) / 2.1 \\
        Aria~\cite{li2024aria} & 128 & 14.1 ($\pm$ 0.3) / 51.5 / 84.4 / ~4.5~ & 13.0 / 44.0 / 82.7 / ~4.2~ & 13.9 / 54.9 / 85.3 / ~4.4~ & ~7.1~ / 34.2 / 81.8 / ~2.3~ & 27.9 / 56.8 / 83.7 / ~8.9~ & 41.5 ($\pm$ 2.2) / 2.1 \\
        Tarsier-34B~\cite{wang2024tarsier} & 16 & 13.5 ($\pm$ 0.2) / 50.8 / 82.1 / 15.1 & 14.7 / 43.9 / 85.5 / 16.4 & 12.4 / 53.7 / 80.4 / 13.8 & ~7.1~ / 38.1 / 84.0 / ~7.9~ & 28.1 / 61.7 / 83.9 / 31.4 & 37.4 ($\pm$ 2.0) / 2.0 \\
        \midrule
        Qwen2-VL-72B~\cite{wang2024qwen2} & 2 fps & 12.2 ($\pm$ 0.2) / 46.8 / 79.0 / ~7.7~ & 12.0 / 42.5 / 79.2 / ~7.6~ & 11.5 / 48.4 / 78.8 / ~7.3~ & ~5.8~ / 28.6 / 77.8 / ~3.7~ & 27.1 / 59.6 / 80.9 / 17.2 & 40.0 ($\pm$ 1.3) / 2.0 \\
        InternVL2-76B~\cite{chen2024far} & 32 & ~7.4~ ($\pm$ 0.2) / 35.6 / 78.9 / ~0.7~ & ~5.8~ / 27.6 / 76.2 / ~0.6~ & ~7.2~ / 38.1 / 80.1 / ~0.7~ & ~4.4~ / 24.4 / 80.0 / ~0.4~ & 23.1 / 55.0 / 78.1 / ~2.3~ & 44.1 ($\pm$ 2.1) / 2.1 \\
        Pixtral-124B~\cite{agrawal2024pixtral} & 16 & 13.0 ($\pm$ 0.3) / 48.3 / 80.5 / ~3.0~ & 13.9 / 44.6 / 80.2 / ~3.2~ & 11.9 / 50.0 / 80.5 / ~2.7~ & ~6.2~ / 28.3 / 81.8 / ~1.4~ & 27.9 / 55.9 / 83.2 / ~6.4~ & 45.4 ($\pm$ 1.9) / 2.2 \\
        \bottomrule
    \end{tabular}
    }
    \caption{Evaluation results on VidCapBench-AE and VDC. ``Acc'', ``Pre'', ``Cov'', and ``Con'' stand for accuracy, precision, coverage, and conciseness, respectively. For better presentation, we have multiplied ``Con'' by 100. ``Score'' is calculated using GPT-4o based on the method in \cite{chai2024auroracap}}
    \label{tab:VidCapBench results}
\end{table*}

\subsubsection{Keypoints Generation}

% After obtaining the videos, we aimed to generate pertinent and diverse question-answer (QA) pairs.  This approach sought to accurately characterize the aspects relevant to text-to-video (T2V) generation while avoiding an overemphasis on lengthy captions due to excessively granular questioning.  
Based on the focused aspects mentioned in Section~\ref{sec:alignment}, we employ GPT-4o~\footnote{Throughout this paper, we employ GPT-4o-20240806 for GPT-4o and Gemini-1.5-Pro-002 for Gemini.} to generate 40 QA pairs for each video, supplemented by 10 additional question-category combinations generated via expert classifiers.
% This resulted in 50 generated QA pairs associated with each video.  
% These QA pairs were not guaranteed to be accurate or consistently evaluable by GPT-4o. 
Subsequently, we generate three different captions for each video using Gemini with varying random seeds.
The first caption is evaluated three times using GPT-4o, while the subsequent two captions are evaluated once each.
Questions exhibiting inconsistent judgments across the three assessments of the first caption, along with those receiving consistently negative evaluations across all five assessments, are flagged as potentially problematic.
These problematic questions, likely due to their ambiguity or lack of clarity, are manually reviewed.
Any factual inaccuracies within the QA pairs are corrected, and the revised pairs are subsequently re-evaluated to determine their suitability.
% Unclear or meaningless QA pairs were discarded. 
Finally, to maintain a balanced distribution of key aspects across the dataset, additional human annotations are performed to address any dimension imbalances introduced by deletions and modifications.

\subsection{QA Pairs Split}
As discussed in Section~\ref{sec:alignment}, the difficulty of evaluating the QA pairs has a significant impact on the reliability of machine evaluation. Therefore, we split the total QA pairs based on their evaluation consistency using the same strategy in Section~\ref{sec:alignment}. In order to achieve a balance between accuracy and efficiency, we identify and segregate QA pairs that fail to receive consistent evaluations within the dimensions of video motion and physical laws, which are more crucial to T2V generation. Consequently, a total of 1,150 QA pairs are classified as \emph{VidCapBench-HE}, which necessitates human intervention for accurate evaluation. The remaining QA pairs are designated as \emph{VidCapBench-AE}, which can be evaluated automatically. Figure~\ref{fig:demo_qa} presents illustrative examples of QA pairs from both categories within VidCapBench. Additional examples and detailed statistics of the QA pairs by dimension are provided in Appendix~\ref{sec:in-depth}.

% \subsection{Subset by Evaluation Stability}

% % We analyze the stability of VidCapBench.
% Five distinct captions were generated for each video using Gemini, following the formats specified by MiraData, Vript, DREAM-1K, HunyuanVideo~\cite{kong2024hunyuanvideo}, and free-form.
% GPT-4o was employed as the judge model, providing three judgments per QA.
% % We first examine evaluation stability, defined as the consistency of the evaluator's judgments. 
% Table~\ref{tab:VidCapBench Stable Evals} summarizes the proportion of stable evaluations (i.e., consistent three judgments) across different dimensions.
% QAs on VA and VC exhibit high stability, while questions related to VM and PL demonstrate considerable instability.
% Subsequently, we probe into those questions consistently judged as wrong and conduct human evaluation.
% Discrepancies between human and automated judgments identified unreliable automated evaluations.  Finally, unstable and unreliably auto-evaluated questions were combined to form a subset requiring human evaluation.

\section{Experiments}

\subsection{Experimental setup}

% \noindent\textbf{Benchmark}. We compare VidCapBench with public dense video captioning benchmarks, specifically DREAM-1K and VDC.
% Experiments are conducted using the official metrics for each benchmark, including Precision/Recall/F1, and Acc/VDCScore.
% To ensure a fair evaluation, each result is averaged over three independent runs, utilizing GPT-4o with random seeds set to 0, 1, and 2, respectively.

\noindent\textbf{Captioning models}.
A variety of vision language models that demonstrate strong captioning capabilities are evaluated.
When available, their official prompts are utilized; otherwise, the generic prompt ``Describe the video in detail'' is employed.
Greedy decoding is applied across all models to minimize the influence of stochasticity.

\noindent\textbf{Environment}.
All experiments are conducted on A800-80GB GPUs using bfloat16 precision.
To ensure a fair evaluation, each result is averaged over three independent runs, utilizing GPT-4o with random seeds set to 0, 1, and 2, respectively.
Frame rate, resolution, and video decoding follow official recommendations where provided.
Otherwise, the ``decord'' library is used to extract 16 frames for captioning, ensuring a minimum frame rate of 1~fps on VidCapBench.
Fifteen experienced annotators, familiar with VidCapBench, provide reliable annotations for the generated captions.

\begin{figure*}[htp]
  \includegraphics[width=\linewidth, height=5.04cm]{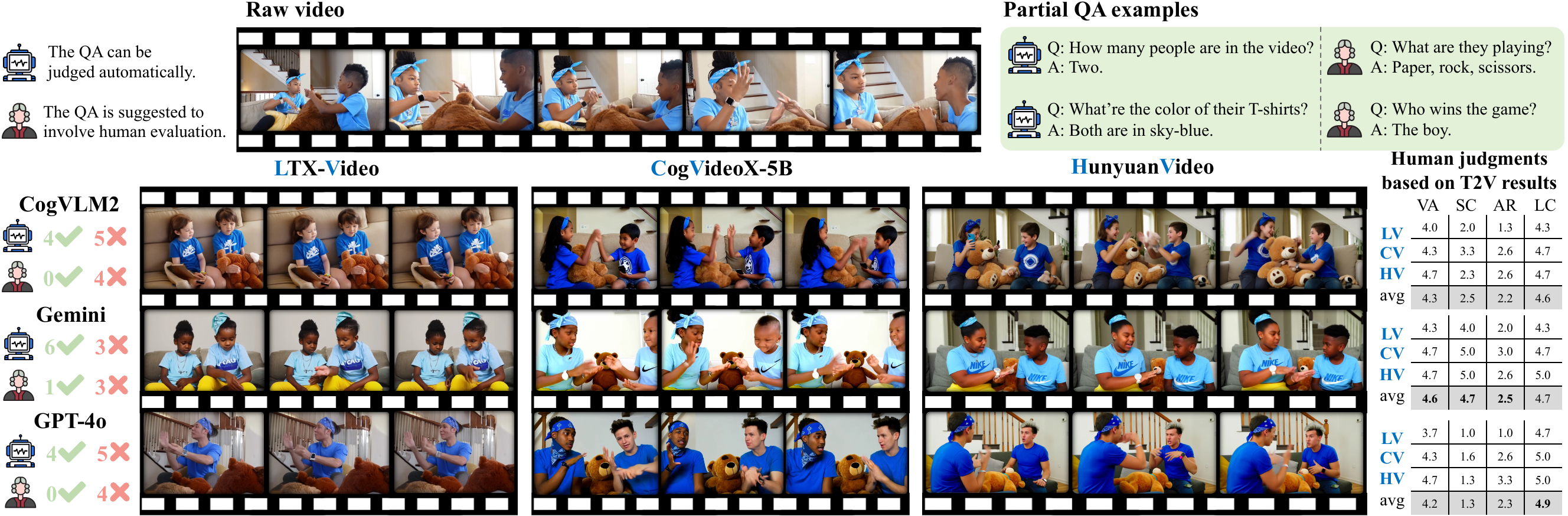}
  \caption{Illustration of the training-free T2V verification for video caption evaluation. ``VA'', ``SC'', ``AR'', and ``LC'' denote the four key dimensions of T2V quality evaluation: ``Visual Aesthetics'', ``Subject Consistency'', ``Action Relevance'', and ``Logical Coherence'', respectively. In this case, the video is associated with nine QA pairs in VidCapBench-AE and four QA pairs in VidCapBench-HE. The similarity between the generated video and the original video, as well as the overall generation quality, are strongly correlated with the evaluation results in VidCapBench. Among the captioning models compared, Gemini exhibits the best performance.}
  \label{fig:casestudy}
\end{figure*}

\subsection{Performance on VidCapBench}

We first analyze model performance on the VidCapBench-AE. As shown in Table~\ref{tab:VidCapBench results}, GPT-4o and Gemini achieve outstanding performance across multiple dimensions and metrics. However, our analysis reveals a notable tendency for GPT-4o, as well as InternVL2-76B, to produce redundant outputs. In contrast, Tarsier-34B generates relatively concise captions, which contributes to its superior Con score, while also maintaining remarkable results in other metrics. Additionally, models specifically trained on large-scale dense captioning data, such as CogVLM2-Caption and Aria, manifest a distinct performance advantage.

The evaluation results also indicate that some models exhibit specialized proficiency in specific dimensions while showing limitations in others. A typical example is Pixtral-12B, which excels in Video Aesthetics and Physical Laws but underperforms in Video Motion and Video Content. While open-source models generally lag behind proprietary models in Video Motion, they exhibit comparable capabilities in Physical Laws. Notably, certain open-source models, particularly Pixtral-12B and Tarsier-34B, even surpass GPT-4o in the Acc score within this dimension.

Regarding evaluation stability, VidCapBench demonstrates superior consistency compared to VDC, which exhibits significant variability across its three evaluation runs. The enhanced stability in VidCapBench contributes to more reliable T2V guidance, establishing it as a more robust evaluation framework for video captioning tasks.

\begin{table}
    \centering
    \resizebox{\columnwidth}{!}{
    \begin{tabular}{l|c c c c c c}
    \toprule
        \multirow{2}{*}{\textbf{Model}} & \multicolumn{3}{c}{\textbf{VidCapBench-AE}} & \multicolumn{3}{c}{\textbf{VidCapBench-HE}} \\
        ~ & Auto & Human & Diff. & Auto & Human & Diff. \\
        \midrule
        GPT-4o & 56.3 & 55.2 & 1.1 & 45.2 & 52.4 & \textbf{~7.2~} \\
        Gemini 1.5 Pro & 53.4 & 52.6 & 0.8 & 55.3 & 49.2 & \textbf{6.1} \\
        Qwen2-VL-72B & 46.7 & 47.4 & 0.7 & 48.5 & 56.8 & \textbf{~8.3~} \\
        CogVLM2-Caption & 47.7 & 48.9 & 1.2 & 51.5 & 56.2 & \textbf{~4.7~} \\
        Tarsier-34B & 52.3 & 51.9 & 0.4 & 57.2 & 51.3 & \textbf{~5.9~} \\ 
        \bottomrule
    \end{tabular}
    }
    \caption{Comparison between automated evaluation and human evaluation on VidCapBench-AE and VidCapBench-HE, respectively. ``Diff.'' stands for the absolute value of the difference between the two evaluation methods.  We choose Pre as the representative here.}
    \label{tab:VidCapBench human eval compare}
\end{table}

\subsection{Fine-grained Analysis of VidCapBench}~\label{sec:fine-grained analysis}
This section focuses on five representative captioning models: GPT-4o, Gemini, Qwen2-VL-72B, CogVLM2-Caption, and Tarsier-34B.
We conduct a fine-grained analysis of their performance to demonstrate the validity of VidCapBench.

\begin{table}
    \centering
    \resizebox{\columnwidth}{!}{
    \begin{tabular}{l|c c c c c c}
    \toprule
        \multirow{2}{*}{\textbf{Model}} & \multirow{2}{*}{\textbf{Prompt Formats}} & \multicolumn{4}{c}{\textbf{Overall}} & \multirow{2}{*}{\makecell{\textbf{Token} \\ \textbf{Num}}} \\
        ~ & ~ & Acc & Pre & Cov & Con & ~ \\
        \toprule
        \multirow{4}{*}{Gemini 1.5 Pro} & MiraData & 17.3 & 57.8 & 87.4 & ~4.6~ & 377.7 \\
        ~ & DREAM-1K & 15.3 & 56.7 & 83.1 & ~9.6~ & 159.2  \\
        ~ & Vript & 17.3 & 57.4 & 88.2 & ~6.4~ & 270.0 \\
        ~ & Hunyuan & 17.0 & 56.9 & 86.4 & ~5.6~ & 305.4 \\
        \midrule
        \multirow{4}{*}{Qwen2-VL-72B} & MiraData & 12.6 & 51.4 & 80.5 & ~4.7~ & 267.6 \\
        ~ & DREAM-1K & 11.4 & 51.4 & 76.0 & ~7.9~ & 144.2 \\
        ~ & Vript & 13.5 & 51.5 & 82.2 & ~6.9~ & 194.7 \\
        ~ & Hunyuan & 13.9 & 52.0 & 81.6 & ~5.3~ & 262.7 \\
        \bottomrule
    \end{tabular}
    }
    \caption{Evaluation comparison between different prompt formats in VidCapBench-AE. Detailed prompts are provided in the Appendix~\ref{sec:prompt_type}.}
    \label{tab:VidCapBench Diff Prompt Formats Compare}
\end{table}

% \begin{figure*}[htp]
%   \includegraphics[width=\linewidth, height=5.1cm]{figures_v2/casestudy.pdf}
%   \caption{Illustration of the training-free T2V verification for video caption evaluation. ``VA'', ``SC'', ``AR'', and ``LC'' denote the four key dimensions of T2V quality evaluation: ``Visual Aesthetics'', ``Subject Consistency'', ``Action Relevance'', and ``Logical Coherence'', respectively. In this case, the video is associated with nine QA pairs in VidCapBench-AE and four QA pairs in VidCapBench-HE. The similarity between the generated video and the original video, as well as the overall generation quality, are strongly correlated with the evaluation results in VidCapBench. Among the captioning models compared, Gemini exhibits the best performance.}
%   \label{fig:casestudy}
% \end{figure*}

\noindent\textbf{Human evaluation consistency}.
To validate the reliability of automatic evaluation on VidCapBench-AE and the necessity of human intervention in VidCapBench-HE, we investigate the discrepancies between automatic and human evaluation on both subsets. Specifically, human annotators are engaged to assess model performance on VidCapBench-HE (detailed results are in Appendix~\ref{sec:discusssion}), and their judgments are compared with those derived from automatic evaluations. Additionally, we randomly sample 1,150 QA pairs from VidCapBench-AE for human annotation and subsequent comparative analysis. The results, summarized in Table~\ref{tab:VidCapBench human eval compare}, reveal strong consistency between human and automatic evaluations for all five models on VidCapBench-AE. Conversely, substantial inconsistencies emerge on VidCapBench-HE. These findings highlight the reliability of automatic evaluation on a stable-evaluation QA subset and the critical role of human annotation in accurately evaluating video captions in certain contexts.
% These findings highlight the critical role of human annotation in accurately evaluating video captions in certain contexts.

\noindent\textbf{Impact of caption format}.
The above experiments do not specify caption formats.
In practice, different T2V models demand distinct caption formats.
Therefore, we evaluate the impact of different caption formats, including MiraData, Vript, HunyuanVideo~\cite{kong2024hunyuanvideo}, and DREAM-1K, on the performance of Gemini and Qwen2-VL-72B.
Table~\ref{tab:VidCapBench Diff Prompt Formats Compare} presents the results using GPT-4o as the judge.
Different formats primarily affect caption length. Shorter captions, such as those in the DREAM-1K format, often lack comprehensive details of videos, resulting in lower Acc and Cov scores. However, the Pre score, which measures the proportion of partially correct answers, remains stable across formats, which underscores the robustness of VidCapBench in evaluating captions of varying formats. 

\begin{table}
    \centering
    \resizebox{\columnwidth}{!}{
    \begin{tabular}{l|cccccc}
    \toprule
        \multirow{2}{*}{\textbf{T2V}} & \multirow{2}{*}{\textbf{Dimensions}} & \multicolumn{3}{c}{\textbf{Captioning Models}} \\
        ~ & ~ & \textbf{GPT-4o} & \textbf{Gemini} & \textbf{CogVLM2} \\
        \toprule
        \multirow{4}{*}{HunyuanVideo} & Visual aesthetics & 3.72 & \textbf{3.84} & 3.60  \\
        ~ & Subject consistency & 3.66 & \textbf{3.70} & 3.04 \\
        ~ & Action relevance & \textbf{3.38} & 3.32 & 2.50 \\
        ~ & Logical coherence & 3.58 & \textbf{3.60} & 3.56 \\
        \midrule
        \multirow{4}{*}{CogVideoX} & Visual aesthetics & 2.86 & \textbf{2.96} & 2.84 \\
        ~ & Subject consistency & 3.56 & \textbf{3.58} & 3.50 \\
        ~ & Action relevance & \textbf{3.26} & 3.24 & 3.20 \\
        ~ & Logical coherence & \textbf{2.92} & 2.86 & 2.88 \\
        \midrule
        \multirow{4}{*}{LTX-Video} & Visual aesthetics & 2.80 & \textbf{2.82} & 2.34 \\
        ~ & Subject consistency & \textbf{2.96} & 2.94 & 2.54 \\
        ~ & Action relevance & \textbf{2.52} & 2.38 & 1.94 \\
        ~ & Logical coherence & \textbf{2.54} & 2.53 & 2.50 \\
        \bottomrule
    \end{tabular}
    }
    \caption {T2V quality evaluations by human across four dimensions.}
    \label{tab:human video eval}
\end{table}

\subsection{Training-free T2V Verification}
Ideally, we would train multiple identical T2V models from scratch using extensive datasets generated by corresponding captioning models and then evaluate the video quality produced by each caption variant. 
However, considering the influence of data distribution and convergence behavior, such a lengthy validation pipeline might not yield clear and focused conclusions.
Therefore, leveraging the high semantic alignment capabilities of advanced T2V models, we adopt a training-free verification approach.
Specifically, we directly feed captions generated towards videos in VidCapBench into three open-source T2V models: CogVideoX-5B~\footnote{The token limit of its text encoder is extended to 400.}, LTX-Video~\footnote{https://huggingface.co/Lightricks/LTX-Video}, and HunyuanVideo.
The captions are generated using GPT-4o, Gemini, Qwen2-VL-72B,  CogVLM2-Caption, and Tarsier-34B. Figure~\ref{fig:casestudy} provides examples of the generated videos.

% \begin{table}[htp]
%     \centering
%     \resizebox{\columnwidth}{!}{
%     \begin{tabular}{l|cccccc}
%     \toprule
%         \multirow{2}{*}{\textbf{T2V}} & \multirow{2}{*}{\textbf{Metrics}} & \multicolumn{5}{c}{\textbf{Captioner}} \\
%         ~ & ~ & \textbf{GPT-4o} & \textbf{Gemini} & \textbf{Qwen2-VL} & \textbf{CogVLM2} & \textbf{Tarsier} \\
%         \toprule
%         \multirow{4}{*}{HunyuanVideo} & CLIP & 0.3324 & \textbf{0.3491} & 0.3453 & 0.3419 & 0.3413 \\
%         ~ & PSNR & 29.061 & 28.675 & 28.887 & \textbf{29.472} & 27.839 \\
%         ~ & SSIM & 0.8957 & 0.8937 & 0.8969 & \textbf{0.9073} & 0.8811 \\
%         ~ & FVD$\downarrow$ & \textbf{4240.6} & 4349.1 & 4377.0 & 4509.7 & 4468.0 \\
%         \midrule
%         \multirow{4}{*}{CogVideoX} & CLIP & 0.3429 & \textbf{0.3536} & 0.3478 & 0.3434 & 0.3447 \\
%         ~ & PSNR & 25.560 & \textbf{27.095} & 26.124 & 23.228 & 26.672 \\
%         ~ & SSIM & 0.7676 & \textbf{0.7879} & 0.7711 & 0.7577 & 0.7572 \\
%         ~ & FVD$\downarrow$ & \textbf{4222.1} & 4224.2 & 4390.5 & 4226.1 & 4225.2 \\
%         \midrule
%         \multirow{4}{*}{LTX-Video} & CLIP & 0.3474 & \textbf{0.3532} & 0.3523 & 0.3334 & 0.3475 \\
%         ~ & PSNR & \textbf{30.613} & 30.235 & 30.158 & 29.862 & 28.929 \\
%         ~ & SSIM & \textbf{0.8805} & 0.8779 & 0.8760 & 0.8726 & 0.8461 \\
%         ~ & FVD$\downarrow$ & \textbf{4253.0} & 4314.2 & 4347.4 & 4327.0 & 4434.9 \\
%         \bottomrule
%     \end{tabular}
%     }
%     \caption {Automated T2V quality evaluations across four dimensions.}
%     \label{tab:auto video eval}
% \end{table}
\begin{figure}
  \includegraphics[width=\linewidth]{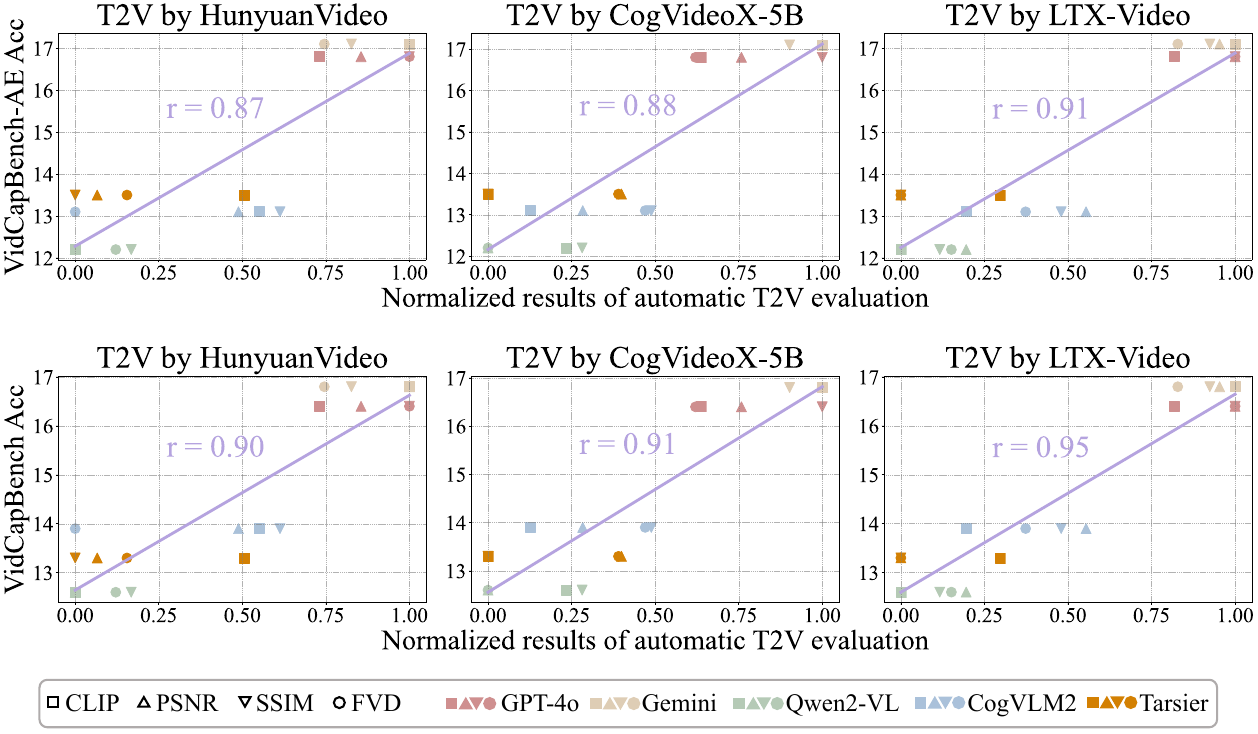}
  \caption{Correlations between automated T2V quality evaluations and VidCapBench-AE Acc (upper) and VidCapBench full set Acc (lower). The Pearson correlation coefficient is denoted by "r".}
  \label{fig:auto T2V eval}
\end{figure}

We conduct automated evaluations on the generated videos across four dimensions:
semantic relevance, quantified using \textit{CLIP} score derived from CLIP-L~\cite{radford2021learning}, which assesses both the model's textual alignment and the caption's suitability for the T2V task; 
aesthetic quality and structural integrity, evaluated using inter-frame \textit{PSNR} and \textit{SSIM};
and fidelity to the original video, measured using \textit{FVD}~\cite{unterthiner2019fvd}.
The correlations between the four T2V evaluation metrics and the Acc on VidCapBench are illustrated in Figure~\ref{fig:auto T2V eval}. The upper panels demonstrate a strong positive correlation between the Acc on VidCapBench-AE and automated T2V quality assessments, with an average Pearson correlation coefficient of 0.89, which substantiates the effectiveness of automated video caption assessments on VidCapBench-AE. Furthermore, the lower panels reveal an enhanced correlation for the Acc on VidCapBench full set, achieving a higher mean Pearson correlation coefficient of 0.92, highlighting the significant impact of human intervention on VidCapBench-HE.
% We observe that, for a given T2V model, the highest video fidelity is achieved by GPT-4o and Gemini, the top performers on VidCapBench, demonstrating the effectiveness of VidCapBench for information transfer.
% Semantic relevance appears to be an inherent model capability, where relatively short captions yield similar CLIP scores for the same T2V model.
% An exception was GPT-4o, which, due to its longer and more redundant captions, consistently produced lower CLIP scores across all three T2V models.

% \begin{figure}
%   \includegraphics[width=\linewidth]{figures_v2/scatter_human.pdf}
%   \caption{Human T2V quality evaluations. The Acc assessed in VidCapBench exhibits a strong relation with human T2V evaluation results. Specifically, the Pearson correlation coefficients for the three T2V models are 0.94, 0.94, and 0.96, respectively.}
%   \label{fig:human T2V eval}
% \end{figure}

We further conduct human evaluations on the videos generated from captions produced by GPT-4o, Gemini, and CogVLM2, scoring aspects in visual aesthetics, subject consistency, action relevance, and logical coherence on a scale of 1 to 5.
Each annotator is assigned approximately 130 sets of videos (9 videos per set), ensuring that each video receives three independent evaluations.
To maintain objectivity and minimize bias, annotators are presented with an original video alongside nine T2V generations in a randomized and anonymized order.
The results, summarized in Table~\ref{tab:human video eval}, reveal a similar pattern to the findings in VidCapBench. Specifically, GPT-4o and Gemini exhibit comparable performance, both significantly outperforming CogVLM2 across the four dimensions, which further validates the alignments between caption evaluations on VidCapBench and the T2V qualities.
% The results, summarized in Table~\ref{tab:human video eval}, reveal a similar pattern with the corresponding Acc and Pre in VidCapBench. Notably, the relative performance ranking of the three captioning models observed in VidCapBench is preserved, with Gemini performing the best in Video Aesthetics and GPT-4o achieving the highest score in Video Motion.

% The results are summarized in Table~\ref{tab:human video eval}. Comparing these results with the corresponding Acc and Pre scores, we observe that the partial ordering of the three captioning models on VidCapBench is preserved in the T2V results, where Gemini performs the best in Video Aesthetics and GPT-4o reaches the best score in Video Motion.
% The normalized evaluation results, shown in Figure~\ref{fig:human T2V eval}, also exhibit a strong correlation with the Acc measured in VidCapBench, achieving an average Pearson correlation coefficient of 0.95 across three T2V models.

The above findings demonstrate that, for a production-ready T2V model, the quality of captions as assessed in VidCapBench is highly correlated with the quality of the generated videos. Consequently, improving caption quality emerges as a crucial strategy for enhancing T2V model performance, regardless of whether training-based or training-free methods are employed.

\section{Conclusion}

VidCapBench introduces a comprehensive evaluation framework for video captioning in T2V generation, assessing across four key dimensions: video aesthetics, video subject, video motion, and physical laws.
To cater to different evaluation needs, VidCapBench comprises two subsets: one designed for automated evaluation prioritizing speed, and the other for human evaluation prioritizing accuracy.
Compared to existing benchmarks, VidCapBench exhibits greater stability and reliability. Furthermore, the strong correlation between scores on VidCapBench and T2V quality metrics demonstrates its potential for guiding T2V training processes.

\section*{Limitations}
While VidCapBench provides a stable and reliable framework for evaluating captioning models in the aspect of T2V generation, it has certain limitations. Specifically, VidCapBench focuses primarily on captioning tasks, thereby excluding the assessment of other model capabilities, such as mathematical reasoning.

\section*{Ethical Considerations}
Regarding the ethical considerations, it is worth noting that some T2V models may generate biased or harmful content, which could perpetuate stereotypes or misinformation. We strongly emphasize the importance of responsible use and encourage developers to implement robust safeguards, including bias detection mechanisms and content moderation systems, to mitigate these risks.
% Regarding the ethical impact, it is important to note that some video segments within our dataset have been sourced from public user-generated content (UGC) platforms that are free from copyright restrictions. However, these materials should not be employed for any commercial purposes.

\section*{Acknowledgements}
This work is jointly sponsored by National Natural Science Foundation of China (62236010, 62141608, 62206291).
\bibliography{custom}

\clearpage
\appendix

\section*{Appendix}

\section{Detailed Settings of Experiments}

In this section, we provide the detailed settings for the experiments.

\subsection{Captioning Models}

In this paper, we select the following models as representatives of mainstream caption technologies.

\begin{itemize}[leftmargin=*]
\item \textbf{Llava-Next-Video}: This model represents a significant advancement within the native Llava family. Its principal advancements encompass the integration of AnyRes and a more diverse dataset, making it a strong representative of the Llava family. It demonstrates impressive zero-shot performance on video understanding tasks.
\item \textbf{LongVA}: This model improves the long context capability via zero-shot transfer from language to vision, which can process 2,000 frames or over 200K visual tokens.
\item \textbf{mPLUG-Owl3}: This approach leverages the cross-attention mechanism to fuse the vision modality and language modality, somewhat like the Flamingo and llama3V architecture.
\item \textbf{InternVL2}: A family of vision language models that consumes a large amount of instruction data.
\item \textbf{Qwen2-VL}: A family of vision language models that employs 3D RoPE and NaViT, getting rid of the resized aspect ratio of the video frames.
% \item \textbf{Llava-Video}: A vision language model trained on fully synthetic data. This model tends to generate a very long context for the video caption task.
\item \textbf{Pixtral}: A family of vision language models that employs 2D RoPE, a prominent representative on 12B-scale and 124B-scale.
\item \textbf{CogVLM2-Caption}: A captioning model linked to CogVideoX, also a typical caption-related SFT model from existing vision language models.
\item \textbf{Aria}: Representative of the MoE-based vision language models.
\item \textbf{Tarsier}: A captioning model that is designed to describe the events in a video.
\end{itemize}

\subsection{Licensing}
The benchmarks and captioning models used in this paper are solely for academic purposes, as permitted by their respective licenses below.

\noindent\textbf{Benchmarks license.}
DREAM-1K and VDC are licensed under the Apache-2.0 License.

\noindent\textbf{Captioning models license.}
Llava-Next-Video, LongVA, mPLUG-Owl3, Qwen2-VL, Pixtral, Aria, CogVLM2-Caption, and Tarsier adopt the Apache-2.0 License. InternVL-2 is under the MIT License.

\subsection{Prompts for Caption Generation}~\label{sec:prompt_type}
The prompts to generate captions for all models are depicted in Figure~\ref{fig:caption_free}-\ref{fig:caption_hunyuan}.
We have verified that all models can follow the instructions and provide the captions with the correct format.

\begin{CJK*}{UTF8}{gbsn}
\begin{figure}[!h]
\begin{tcolorbox}[colback=white!95!gray,colframe=gray!50!black,rounded corners,label={mid-zh-prompts-1}, title={Free-form caption}]
Describe the video in detail.
\end{tcolorbox}
\caption{Prompts to generate free-form caption.}\label{fig:caption_free}
\end{figure}
\end{CJK*}

\begin{CJK*}{UTF8}{gbsn}
\begin{figure}[!h]
\begin{tcolorbox}[colback=white!95!gray,colframe=gray!50!black,rounded corners,label={mid-zh-prompts-2}, title={MiraData format}]
Carefully look at all frames and then generate a faithful description about the video of the following content.\\
1. Short caption that briefly describes the main subject and actions in the video.\\
2. Dense caption that covers the main subject, movements, style, backgrounds, and cameras.\\
3. Main Object that describes the primary object or subject in the video, capturing their attributes, actions, positions, and movements.\\
4. Background that provides context about the environment or setting, including objects, location, weather, and time.\\
5. Camera Movements that detail any camera pans, zooms, or other movements.\\
6. Video Style: covers the artistic style, as well as the visual and photographic features of the video.\\
No need to provide summary content. Do not describe each frame individually. Do not reply with words like `first frame'. Please provide the description strictly in the following format: `1. Short Caption: ... \\
2. Dense Caption: ... \\
3. Main Object Caption: ... \\
4. Background Caption: ... \\
5. Camera Caption: ... \\
6. Style Caption: ...'. The description should be useful to re-generate the video.
\end{tcolorbox}
\caption{Prompts to generate captions in MiraData-required format.}\label{fig:caption_miradata}
\end{figure}
\end{CJK*}

\begin{CJK*}{UTF8}{gbsn}
\begin{figure}[!h]
\begin{tcolorbox}[colback=white!95!gray,colframe=gray!50!black,rounded corners,label={mid-zh-prompts-3}, title={DREAM-1K format}]
You are to assist me in accomplishing a task about the input video. Reply to me with a precise yet detailed response. For how you would succeed in the recaptioning task, read the following Instructions section and Then, make your response with an elaborate paragraph.\\
\# Instructions\\
1. Avoid providing over detailed information such as color, and counts of any objects as you are terrible regarding observing these details\\
2. Instead, you should carefully go over the provided video and reason about key information about the overall video\\
3. If you are not sure about something, do not include it in your response.\\
\# Task\\
Describe the background, characters, and actions in the provided video.
\end{tcolorbox}
\caption{Prompts to generate captions in DREAM-1K-required format.}\label{fig:caption_dream}
\end{figure}
\end{CJK*}

\begin{CJK*}{UTF8}{gbsn}
\begin{figure}[!h]
\begin{tcolorbox}[colback=white!95!gray,colframe=gray!50!black,rounded corners,label={mid-zh-prompts-4}, title={Vript format}]
Based on the successive frames from the video, please describe:\\
1) the shot type (15 words)\\
2) the camera movement (15 words)\\
3) what is happening as detailed as possible (e.g. plots, characters’ actions, environment, light, all objects, what they look like, colors, style, etc.) (150 words)\\
4) Summarize the content to title the scene (10 words)\\
Do not describe the frames individually but the whole clip.
\end{tcolorbox}
\caption{Prompts to generate captions in Vript-required format.}\label{fig:caption_vript}
\end{figure}
\end{CJK*}

\begin{CJK*}{UTF8}{gbsn}
\begin{figure}[!h]
\begin{tcolorbox}[colback=white!95!gray,colframe=gray!50!black,rounded corners,label={mid-zh-prompts-5}, title={Hunyuan format}]
Carefully look at all frames and then generate a faithful description about the video of the following content.\\
1. Short description capturing the main content of the scene.\\
2. Dense description detailing the scene’s content, which notably includes scene transitions and camera movements that are integrated with the visual content, such as camera following some subject.\\
3. Background describing the environment in which the subject is situated.\\
4. Style characterizing the style of the video, such as documentary, cinematic, realistic, or sci-fi.\\
5. Shot type identifying the type of video shot that highlights or emphasizes specific visual content, such as aerial shot, close-up shot, medium shot, or long shot.\\
6. Lighting describing the lighting conditions of the video.\\
7. Atmosphere conveying the atmosphere of the video, such as cozy, tense, or mysterious.\\
No need to provide summary content. Do not describe each frame individually. Do not reply with words like `first frame'. Please provide the description strictly in the following format: `1. Short Description: ...\\
2. Dense Description: ...\\
3. Background: ...\\
4. Style: ...\\
5. Shot Type: ...\\
6. Lighting: ...\\
7. Atmosphere: ...'. The description should be useful for AI to re-generate the video.
\end{tcolorbox}
\caption{Prompts to generate captions in Hunyuan-required format.}\label{fig:caption_hunyuan}
\end{figure}
\end{CJK*}

% \newpage
% \clearpage

\subsection{Prompts for Judgment}
The prompts for the judge model is listed in Figure~\ref{fig:prompt_answer} and Figure~\ref{fig:prompt_judge}.

\begin{table*}[!h]
    \centering
    \resizebox{\linewidth}{!}{
    \begin{tabular}{lc|c|cccccc}
    \toprule
        \multirow{3}{*}{\textbf{Model}} & \multirow{3}{*}{\textbf{Frames}} & \textbf{DREAM-1K} & \multicolumn{6}{c}{\textbf{VDC}} \\
        ~ & ~ & \textbf{Overall} & \textbf{Overall} & \textbf{Detailed} & \textbf{Camera} & \textbf{Short} & \textbf{Background} & \textbf{Object} \\
        ~ & ~ & \textbf{F1 / P / R} & \textbf{Acc / Score} & \textbf{Acc / Score} & \textbf{Acc / Score} & \textbf{Acc / Score} & \textbf{Acc / Score} & \textbf{Acc / Score} \\
        \midrule
        GPT-4o-20240806 & 16 & 37.8 ($\pm$ 0.6) / 37.9 / 37.8 & 46.1 ($\pm$ 1.0) / 2.2 & 50.1 / 2.3 & 52.5 / 2.4 & 34.3 / 1.9 & 44.6 / 2.1 & 48.8 / 2.3 \\
        Gemini-1.5-Pro-002 & - & 37.7 ($\pm$ 0.3) / 37.1 / 38.3 & 40.5 ($\pm$ 1.6) / 2.0 & 46.3 / 2.2 & 40.9 / 2.1 & 30.0 / 1.7 & 41.2 / 2.0 & 44.3 / 2.1 \\
        \midrule
        Llava-Next-Video-7B & 16 & 23.4 ($\pm$ 1.2) / 27.7 / 20.3 & 37.6 ($\pm$ 2.2) / 1.9 & 38.8 / 2.0 & 46.2 / 2.2 & 29.5 / 1.8 & 38.4 / 2.0 & 35.4 / 1.9 \\
        LongVA-7B & 128 & 24.7 ($\pm$ 1.5) / 29.8 / 21.1 & 36.1 ($\pm$ 1.9) / 1.9 & 38.7 / 2.0 & 44.0 /2.1 & 22.0 / 1.5 & 36.4 / 1.9 & 39.7 / 2.0 \\
        mPLUG-Owl3-7B & 16 & 27.9 ($\pm$ 1.0) / 30.2 / 26.0 & 36.5 ($\pm$ 2.7) / 1.9 & 35.3 / 1.9 & 43.9 / 2.1 & 25.0 / 1.6 & 39.0 / 2.0 & 39.4 / 2.0 \\
        InternVL2-8B & 32 & 27.9 ($\pm$ 0.8) / 28.0 / 27.9 & 37.4 ($\pm$ 2.4) / 1.9 & 41.1 / 2.1 & 43.8 / 2.1 & 24.4 / 1.6 & 37.7 / 1.9 & 40.2 / 2.0 \\
        Qwen2-VL-7B & 2 fps & 29.8 ($\pm$ 0.9) / 34.5 / 26.2 & 39.6 ($\pm$ 1.6) / 2.0 & 42.6 / 2.1 & 46.4 / 2.2 & 29.7 / 1.7 & 38.1 / 1.9 & 40.9 / 2.0 \\
        % Llava-Video-7B & 32 & 35.6 ($\pm$ 0.7) / 35.1 / 36.1 & 43.4 ($\pm$ 1.8) / 2.1 & 47.9 / 2.3 & 46.1 / 2.2 & 31.9 / 1.8 & 44.3 / 2.1 & 46.8 / 2.2 \\
        \midrule
        Pixtral-12B & 16 & 20.9 ($\pm$ 1.7) / 26.3 / 17.3 & 39.0 ($\pm$ 2.3) / 2.0 & 43.0 / 2.1 & 47.2 / 2.2 & 29.6 / 1.7 & 37.6 / 1.9 & 37.6 / 1.9 \\
        CogVLM2-Caption & 1 fps & 28.9 ($\pm$ 0.3) / 29.2 / 28.6 & 42.8 ($\pm$ 1.4) / 2.1 & 44.7 / 2.2 & 47.7 / 2.2 & 30.5 / 1.7 & 45.3 / 2.2 & 45.8 / 2.2 \\
        Aria & 128 & 31.5 ($\pm$ 0.9) / 31.3 / 31.7 & 41.5 ($\pm$ 2.2) / 2.1 & 48.0 / 2.3 & 44.7 / 2.2 & 30.1 / 1.7 & 38.8 / 2.0 & 45.7 / 2.2 \\
        Tarsier-34B & 16 & 38.7 ($\pm$ 0.4) / 45.0 / 33.9 & 37.4 ($\pm$ 2.0) / 2.0 & 40.3 / 2.1 & 43.1 / 2.1 & 25.0 / 1.6 & 39.5 / 2.0 & 39.3 / 2.0 \\
        \midrule
        Qwen2-VL-72B & 2 fps & 30.8 ($\pm$ 0.8) / 34.7 / 27.7 & 40.0 ($\pm$ 1.3) / 2.0 & 43.6 / 2.1 & 46.9 / 2.2 & 28.0 / 1.7 & 39.9 / 2.0 & 41.7 / 2.1 \\
        InternVL2-76B & 32 & 25.4 ($\pm$ 1.2) / 27.1 / 23.9 & 44.1 ($\pm$ 2.1) / 2.1 & 48.7 / 2.3 & 52.7 / 2.3 & 23.6 / 1.6 & 47.5 / 2.2 & 47.8 / 2.2 \\
        Pixtral-124B & 16 & 30.0 ($\pm$ 0.7) / 30.5 / 29.6 & 45.4 ($\pm$ 1.9) / 2.2 & 48.6 / 2.2 & 51.6 / 2.3 & 34.1 / 1.9 & 47.0 / 2.2 & 45.6 / 2.2 \\
        \bottomrule
    \end{tabular}
    }
    \caption{Evaluation results on DREAM-1K and VDC benchmark. Regrading DREAM-1K, ``F1'', ``P'', and ``R'' stand for F1 score, precision, and recall respectively. Regarding VDC, ``Acc'' stands for accuracy, ``Score'' is calculated using GPT-4o based on the method in \cite{chai2024auroracap}}
    \label{tab:baseline_result}
\end{table*}

\FloatBarrier

\begin{CJK*}{UTF8}{gbsn}
\begin{figure}[h]
\begin{tcolorbox}[colback=white!95!gray,colframe=gray!50!black,rounded corners,label={mid-zh-prompts-6}, title={Prompts to answer.}]
You are an intelligent chatbot to answer questions given a detailed description of a video or image.\\
Your answer should be a short sentence or phrase.\\
\\
Description:\\
\{caption\}\\
\\
Question: \{question\}
\end{tcolorbox}
\caption{Prompts to answer the questions by the judge model.}\label{fig:prompt_answer}
\end{figure}
\end{CJK*}

\begin{CJK*}{UTF8}{gbsn}
\begin{figure*}[hp]
\begin{tcolorbox}[colback=white!95!gray,colframe=gray!50!black,rounded corners,label={mid-zh-prompts-7}, title={Prompts to judge.}]
Please act as an impartial and objective judge and evaluate the correctness of generative outputs for question-answer pairs provided by a Large Language Model.\\
Your evaluation should be mainly based on whether the predicted answer mentions the provided correct answer comprehensively and accurately.\\
You need to first comprehensively understand the content of the origin QA pairs and grasp the content of it. Then, you need to analyze if it is accurately reflected in the predicted answer generated by the Large Language Model. For each predicted answer, provide a brief analysis explaining your reasoning for the score.\\

You will then select from the following options to score the degree to which the model-predicted answer reflects the correct answer:\\
- Score: 2, the predicted answer comprehensively and accurately reflects the content of the correct answer.\\
- Score: 1, the predicted answer mentions the correct answer, but the information is not precise or complete. However, there is no contradiction with the correct answer.\\
- Score: 0, the predicted answer does not mention the correct answer at all.\\
- Score: -1, the predicted answer contradicts the correct answer or has a partial misrepresentation.\\
Requirements:\\
(1) If the predicted answer mentions a subject that is not mentioned in any of the correct answer, and upon reasoning, it is possible that this subject is misidentified from a subject in the correct answer, then prioritize handling it as -1.\\
(2) When scoring, if the subject in the correct answers is a specific entity, prioritize scoring based on whether the subject is mentioned or if there is a conflict about the subject. Provided that the description of the subject is accurate, then score based on the accuracy of the attributes, states, or actions.\\
(3) If the correct answer doesn't mention a specific entity, instead, it uses pronouns like ``it'' or ``the person'' to refer to subjects, scoring based solely on the accuracy of the attribute, states, or actions.\\
(4) For color attributes, if the caption describes the color of a subject with a specific word that is different from the word used in the key point, but the two colors are similar, then score it as 1.\\
(5) Please present the result in a JSON dict format:\{"score": score, "analysis": analysis\}.\\
\\
Please help me evaluate whether the predicted answer accurately reflects the correct answer.\\
\\
Question: \{question\}\\
Correct Answer: \{answer\}\\
Predicted Answer: \{prediction\}\\
\\
Your Result:
\end{tcolorbox}
\caption{Prompts to judge the answers by the judge model.}\label{fig:prompt_judge}
\end{figure*}
\end{CJK*}

\subsection{Details of Human Annotators}
Fifteen experienced human annotators, all fluent in English and based in Asia, are recruited from a crowdsourcing platform to participate in the annotation and validation process for VidCapBench over a two-week period. To ensure the quality and reliability of the annotations, we compensate annotators based on the time they spend rather than the number of samples completed, preventing them from rushing through tasks. Annotators are compensated at a rate of 10 USD per hour for both the annotation and evaluation processes.

\section{Analysis on VDC and DREAM-1K}

This section provides the analysis on VDC and DREAM-1K, which we found to be comprehensive but not stable.

\subsection{Model Performance}
We examine the performance of various models on the VDC and DREAM-1K benchmarks, summarized in Table~\ref{tab:baseline_result}. VDC demonstrates significant instability across its three evaluation runs, decreasing the reliability of T2V guidance. While repeated evaluations could mitigate uncertainty, the large volume of QA pairs in VDC imposes substantial constraints on both time and computational resources. Moreover, the VDCScore is quite close among different models, further complicating the accurate assessment of model performance. DREAM-1K, on the other hand, primarily focuses on event-centric descriptions without analysis of other crucial aspects, categorized solely by video type. Similar to the VDC benchmark, DREAM-1K also exhibits significant instability across its three evaluation runs.

\subsection{Stability of VDC}
\begin{table}[hp]
    \centering
    \scalebox{0.85}{
    \begin{tabular}{c c c}
    \toprule
        \textbf{Consistent Times} & \textbf{\# Samples} & \textbf{Proportion} \\
        \midrule
        5 & 39,897 & 41\% \\
        4 & 27,046 & 28\% \\
        3 & 17,479 & 18\% \\
        2 & 8,735  & 9\% \\
        1 & 3,038  & 3\% \\
        0 & 630    & 1\% \\
        \midrule
        Total & 96,825 & 100\% \\
        \bottomrule
    \end{tabular}
    }
    \caption{Evaluation agreements of QA pairs in VDC.}
    \label{tab:VDC_Agreements}
\end{table}
Table~\ref{tab:VDC_Agreements} presents the stability of caption judgment for five vision language models (GPT-4o, Gemini, Qwen2-VL-72B, CogVLM2-Caption, Tarsier-34B), with each caption assessed three times.
Only 41\% of the QA pairs show complete agreement across all five models.  Furthermore, a mere 13\% achieved agreement between at most two models.
These findings highlight the substantial uncertainty inherent in VDC evaluation, a common challenge in the QA-based assessment paradigm.

\begin{CJK*}{UTF8}{gbsn}
\begin{figure}[!h]
\begin{tcolorbox}[colback=white!95!gray,colframe=gray!50!black,rounded corners,label={mid-zh-prompts-8}, title={Unstable case in VDC}]
Video: Z3C2mKVwFAE \\
Question: What \\
Answer: A vibrant \\
Question: Who \\
Answer: The main character \\
Question: What \\
Answer: Twist and turns
\end{tcolorbox}
\caption{Unstable case in VDC: Unclear questions. These are common problems in the VDC benchmark.}\label{fig:unstable_VDC_1}
\end{figure}
\end{CJK*}

\begin{CJK*}{UTF8}{gbsn}
\begin{figure}[!h]
\begin{tcolorbox}[colback=white!95!gray,colframe=gray!50!black,rounded corners,label={mid-zh-prompts-9}, title={Unstable case in VDC}]
Question: How long does the baker spend on dough preparation? \\
Answer: Unspecified \\
Question: What takes place next in the process? \\
Answer: Unspecified \\
Question: Who or what is in the drivers' seat \\
Answer: in the vehicles being observed?, Unspecified \\
Question: Where is the sunlight in the video? \\
Answer: Unspecified, but above the ocean view
\end{tcolorbox}
\caption{Unstable case in VDC: Leading questions. These are common problems in the VDC benchmark.}\label{fig:unstable_VDC_2}
\end{figure}
\end{CJK*}

\subsection{Unstable Cases in VDC}
We probe into the unstable QA pairs in the VDC benchmark.
We carefully analyzed the cases that were labeled ``unstable'', and found that the unclear questions and the leading questions are two common problems, as demonstrated by Figure~\ref{fig:unstable_VDC_1} and Figure~\ref{fig:unstable_VDC_2}, respectively.
Meanwhile, there are also many questions that are difficult to answer, or difficult to locate the direction of answer.
We believe that these forms of questions are not appropriate to involve judge models for judgment.

\section{In-depth Study of VidCapBench}~\label{sec:in-depth}
In this section, we attempt to provide some statistics and visualization of VidCapBench.

\subsection{Detailed Dimensions of Evaluation}
\label{sec:detail_dimensions}
The focused dimensions in VidCapBench are defined as follows.

\noindent\textbf{Video aesthetics} (VA) encompasses the artistic and technical aspects of video creation, from filming techniques to post-production, which includes:
\textit{Composition} \--  Arrangement of objects and characters within the frame;
\textit{Color} \-- Mainly temperature and saturation;
\textit{Lighting} \-- Either natural or artificial lighting;
\textit{Cinematography} \-- Regarding lenses and camera movements;
% Different lenses (e.g., wide-angle, telephoto) and camera movements (e.g., handheld, tracking shots) to influence scene perception and control narrative flow.
% Basic cinematography includes adjusting focal length, camera angles and direction, and movement control.
\textit{Style} \-- Focusing on visual presentation and narrative techniques.
% The unique characteristics of a video in terms of visual presentation, sound design, editing rhythm, and narrative techniques.
% These elements combine to create a consistent style that influences audience perception and emotional response. Common video styles include documentary, cinematic, commercial, artistic, retro, animation, and science fiction.

\noindent\textbf{Video content} (VC) refers to the narrative content presented in the video.
% , typically comprising subjects, backgrounds, and events. Accurate description of video content significantly impacts the semantic consistency of text-to-video generation：
\textit{Subjects} \-- The primary person(s) or object(s) of focus within the frame, including characteristics, attributes, relationships, positions, and poses.
\textit{Background} \-- The non-focal elements of the video, providing visual support and spatial context. 
% Captions should describe the background environment, content, and scene information, including features, attributes, location, objects, position, weather, and time.
\textit{Events} \-- Specific activities or plot points that drive the narrative.
% Understanding and describing these events is crucial for effective video captioning.

\noindent\textbf{Video motion} (VM) encompasses all movement and motions, including:
\textit{Body movements} \-- Dynamic activities performed by subjects, reflecting posture, interactions with the environment, or other subjects.
\textit{Object motion} \-- Changes in object position over time;
\textit{Attribute changes} \-- Alterations in physical form, chemical properties, or motion state of objects (e.g., explosions, ripples, shattering);
\textit{Environmental motion} \-- Movement within the background (e.g., natural phenomena, water movement);
\textit{Spatiotemporal transformations} \-- Techniques that alter the perception of time and space (e.g., slow motion, time-lapse).

\noindent\textbf{Physical laws} (PL) allow for more realistic or dramatic visual expression, even though creativity can somewhat bend them.
% Captions should acknowledge adherence to or deviations from expected physical realities, including gravity, light behavior, and the passage of time.
Specifically, VidCapBench focuses:
\textit{Counter-intuitive scenarios} \--  Identifying and describing scenarios that defy typical physical expectations;
\textit{Geometry} \--  Understanding and describing the spatial relationships between objects and the impact of camera perspective.
\textit{Common Sense} \-- Applying everyday knowledge and intuition to interpret events in a scene.

\begin{figure*}[!h]
  \includegraphics[width=\linewidth, height=10.38cm]{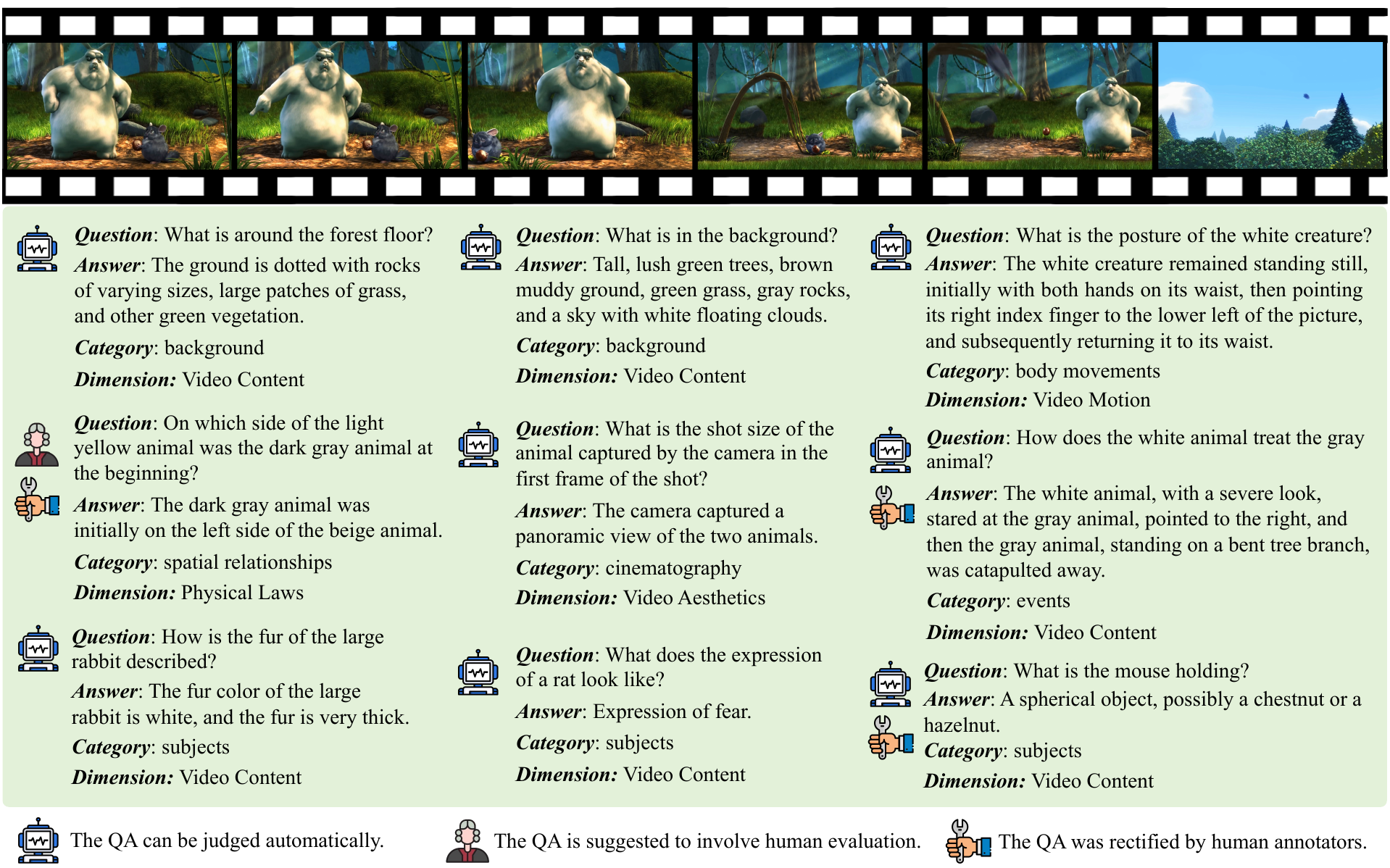}
  \caption{An example of the QA pairs for a video in VidCapBench.}
  \label{fig:demo_qa_6}
\end{figure*}

\begin{figure*}[!h]
  \includegraphics[width=\linewidth, height=11cm]{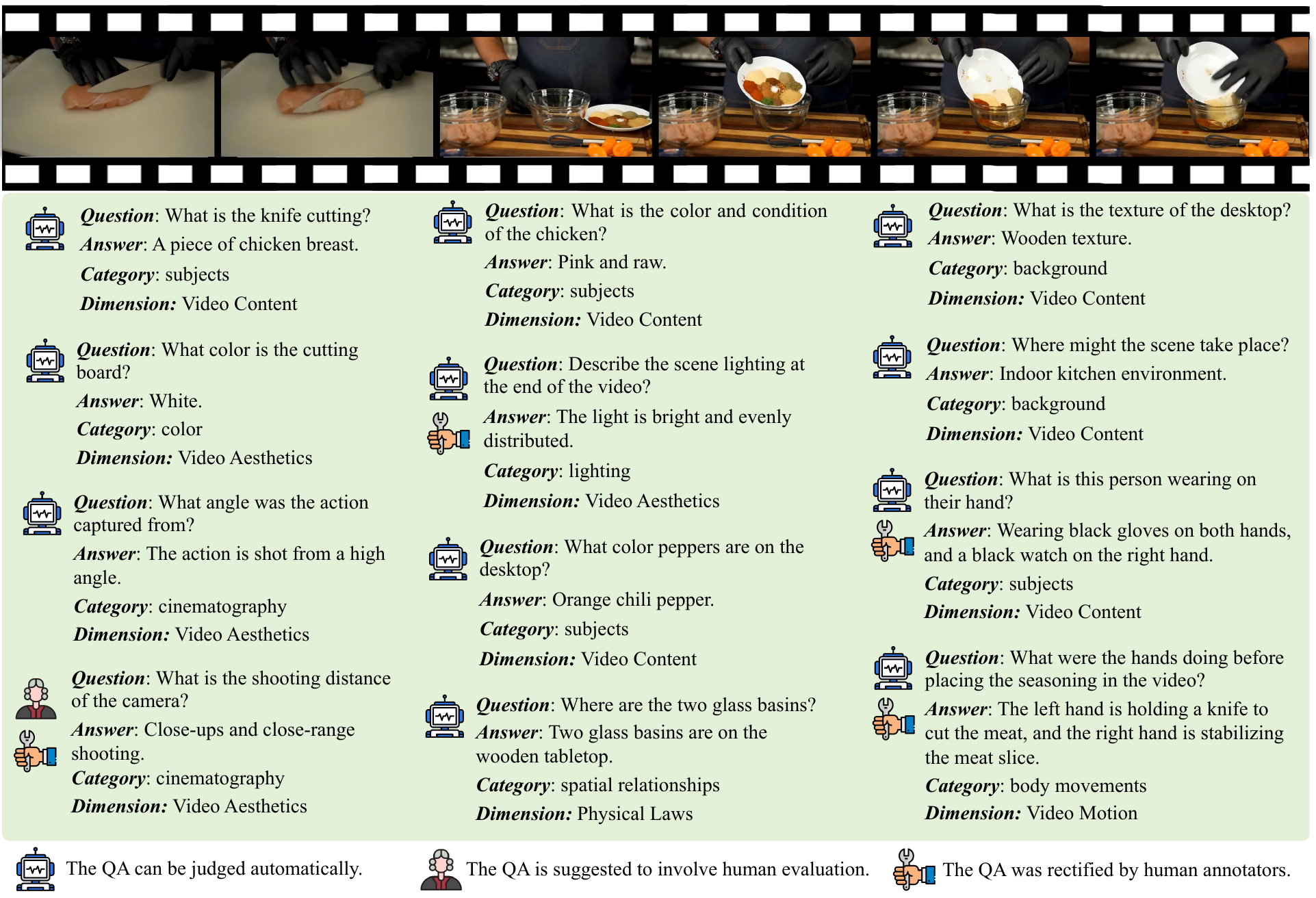}
  \caption{An example of the QA pairs for a video in VidCapBench.}
  \label{fig:demo_qa_7}
\end{figure*}

\label{sec:statistics}
\begin{figure*}[!h]
  \includegraphics[width=\linewidth]{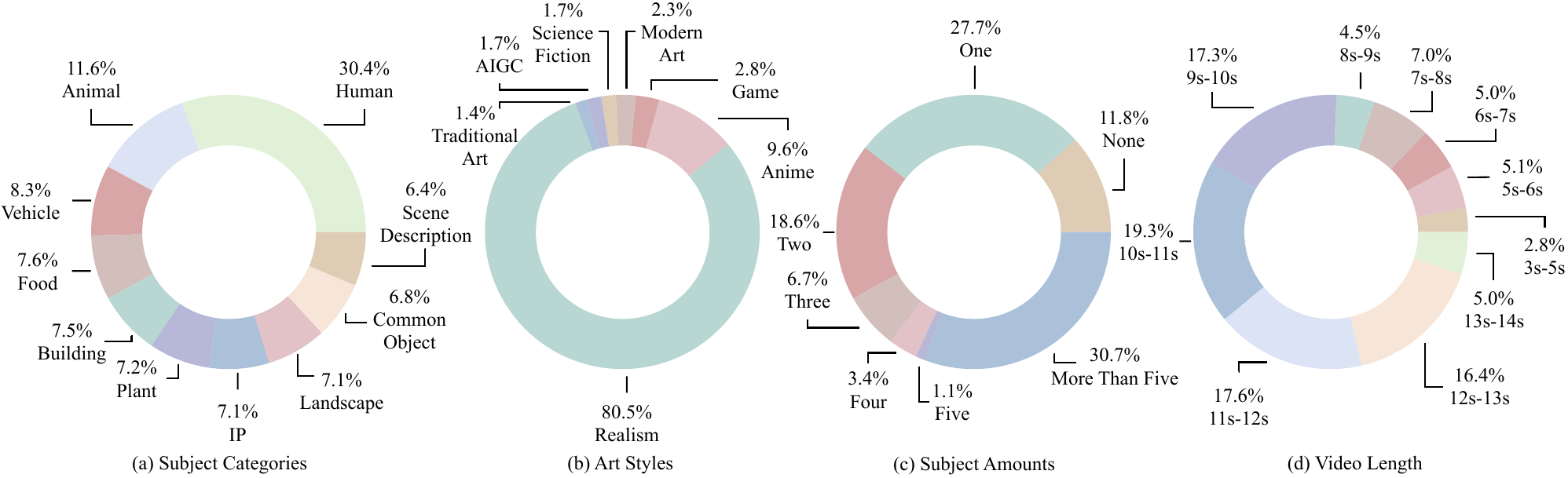}
  \caption{Distribution of videos in VidCapBench on various taxonomies.}
  \label{fig:videostats}
\end{figure*}

\begin{table*}[!h]
    \centering
    \resizebox{\linewidth}{!}{
    \begin{tabular}{c | c c | c c c c | c c c}
    \toprule
        \multirow{2}{*}{\textbf{Data Source}} & \multicolumn{2}{c|}{\textbf{Orientation}} & \multicolumn{4}{c|}{\textbf{Resolution}} & \multicolumn{3}{c}{\textbf{Motion Type}} \\
        ~ & \textbf{Landscape} & \textbf{Portrait} & \textbf{480P} & \textbf{720P} & \textbf{1K} & \textbf{2K} & \textbf{Subject} & \textbf{Camera} & \textbf{Environment} \\
        \midrule
        open-source & 311 & 0 & 9 & 92 & 209 & 1 & 262 & 197 & 47 \\
        YouTube & 116 & 29 & 25 & 98 & 20 & 2 & 134 & 59 & 24 \\
        UGC & 42 & 148 & 6 & 34 & 128 & 22 & 137 & 102 & 38 \\
        \bottomrule
    \end{tabular}
    }
    \caption {Distribution of video contents in VidCapBench. Note that each video may contain multiple motion types.}
    \label{tab:statistics_by_source}
\end{table*}

\subsection{Additional QA Pairs Examples}
Based on the four dimensions outlined in Section~\ref{sec:detail_dimensions}, we have constructed a comprehensive set of QA pairs to assess the quality of captions in the context of T2V generation. In addition to the examples presented in Figure \ref{fig:demo_qa}, we provide further examples in Figures~\ref{fig:demo_qa_6} and \ref{fig:demo_qa_7} to offer a more complete understanding of VidCapBench.

\subsection{Statistics by Dimension}

Figure~\ref{fig:pipeline}(c) depicts the distribution of QA pairs in VidCapBench across dimensions.
We further present statistics of the videos and the annotated keypoints in VidCapBench.
Figure~\ref{fig:videostats}(a) illustrates the subject distribution.
Given the prominence of humans and animals as subjects in T2V generation, we incorporated data with humans as the primary subject in 30\% of instances and animals in 11\%, with the remaining categories evenly distributed. 
Figure~\ref{fig:videostats}(b) presents the distribution of video styles: 80\% real-world footage, 10\% animation/anime, and the remaining 10\% evenly allocated to other artistic styles.
Figure~\ref{fig:videostats}(c) depicts the distribution of the number of subjects. We included 10\% of videos without explicit subjects, 45\% with 1-2 subjects, and 45\% with 3 or more.
Figure~\ref{fig:videostats}(d) shows the video duration distribution, with 40\% of videos being under 10 seconds.

\subsection{Tools in Data Pipeline}

\begin{itemize}[leftmargin=*]
\item Pose Estimation: We use sapiens-pose-1b~\cite{khirodkar2025sapiens} to detect human presence and pose variations within the sampled frames.
\item Object Detection and Grounding: We apply GroundingDINO-base~\cite{liu2025grounding} to the sampled frames based on initial classification keywords, ensuring the target subject appeared in at least 10 frames with sufficient relative size, and record the number of detected objects.
\item Object Tracking: Using bytetrack\_x\_mot20 \cite{zhang2022bytetrack} to track object motion, we label objects as static or consistently trackable.
\item Image Segmentation: We use sam2.1-hiera-large~\cite{ravi2024sam} for image segmentation.  Considering practical T2V generation requirements, we record the number of segmented regions to filter excessively complex scenes (retaining 16-48 segments in practical applications).
\item Optical Character Recognition: We use a lightweight OCR model, namely GOT-OCR-2.0~\cite{wei2024general}, to detect text within the frames.  We aim to minimize the presence of text, particularly subtitles, prioritizing naturally occurring characters.
\end{itemize}

\begin{table*}[!h]
    \centering
    \resizebox{\linewidth}{!}{
    \begin{tabular}{l|c c c c c c}
    \toprule
        \multirow{2}{*}{\textbf{Model}} & \multirow{2}{*}{\textbf{Evaluator}} & \textbf{Overall} & \textbf{Video Aesthetics} & \textbf{Video Content} & \textbf{Video Motion} & \textbf{Physical Laws} \\
        ~ & ~ & \textbf{Acc / Pre / Cov / Con} & \textbf{Acc / Pre / Cov / Con} & \textbf{Acc / Pre / Cov / Con} & \textbf{Acc / Pre / Cov / Con} & \textbf{Acc / Pre / Cov / Con} \\
        \midrule
        \multirow{3}{*}{Gemini 1.5 Pro} & GPT-4o & 17.1 / 54.8 / 87.4 / ~9.2~ & 16.4 / 47.6 / 85.4 / ~8.8~ & 16.9 / 57.8 / 88.5 / ~9.1~ & ~9.8~ / 45.1 / 80.9 / ~5.3~ & 28.4 / 59.3 / 88.2 / 15.3 \\
        ~ & Llama-3.3-70B & 16.7 / 60.2 / 85.2 / ~9.0~ & 18.3 / 53.2 / 91.7 / ~9.8~ & 15.1 / 71.3 / 82.5 / ~8.1~ & 11.3 / 63.1 / 73.2 / ~6.1~ & 26.8 / 70.3 / 76.2 / 14.4 \\
        ~ & Qwen2-72B & 21.9 / 93.4 / 53.2 / 11.7 & 11.6 / 87.4 / 29.7 / ~6.2~ & 24.5 / 95.3 / 62.1 / 13.1 & 12.4 / 83.6 / 48.9 / ~6.6~ & 35.2 / 91.7 / 54.5 / 18.8 \\
        \midrule
        \multirow{3}{*}{Qwen2-VL-72B} & GPT-4o & 12.2 / 46.8 / 79.0 / ~7.7~ & 12.0 / 42.5 / 79.2 / ~7.6~ & 11.5 / 48.4 / 78.8 / ~7.3~ & ~5.8~ / 28.6 / 77.8 / ~3.7~ & 27.1 / 59.6 / 80.9 / 17.2 \\
        ~ & Llama-3.3-70B & 12.7 / 50.9 / 77.1 / ~8.0~ & 14.8 / 49.3 / 84.1 / ~9.4~ & ~9.7~ / 62.1 / 72.9 / ~6.2~ & ~7.2~ / 45.3 / 71.7 / ~4.6~ & 25.4 / 70.1 / 67.6 / 16.1 \\
        ~ & Qwen2-72B & 15.5 / 89.7 / 40.2 / ~9.9~ & ~8.8~ / 88.1 / 24.0 / ~5.6~ & 16.1 / 90.5 / 45.4 / 10.3 & ~8.9~ / 76.2 / 37.3 / ~5.7~ & 31.4 / 92.6 / 47.7 / 20.0 \\
        \bottomrule
    \end{tabular}
    }
    \caption{Evaluation comparison between different evaluators on VidCapBench-AE.}
    \label{tab:VidCapBench Diff Judge Compare}
\end{table*}

\subsection{Statistics by Video Source}

Table~\ref{tab:statistics_by_source} presents statistics for videos from three sources within VidCapBench. To ensure diversity, we purposefully select videos with varying screen orientation, resolution, and motion types, thereby encompassing a wide spectrum of video characteristics and content typologies. Specifically, open-source videos (sourced from ActivityNet, DREAM-1K, NExT-QA, MovieStory101, and Vript-HAL) primarily comprise well-defined and single-event scenarios. Videos from YouTube consist mainly of edited videos, exhibiting greater stylistic diversity. UGC videos, collected from short-video platforms, predominantly reflect everyday life.

% \subsection{Example of QAs}~\ref{sec:demo_qa}

% Figure~\ref{fig:demo_qa} illustrates an example in VidCapBench. In this case, 12 QAs are attached to the video, where two of them are suggested to be judged via human evaluation. We found that the question in this case on the common sense leads to the constant-negative judgment by the judge model, while the question on the events could not get a stable judgment. Besides, four QAs have been rectified by human annotators to improve the stability and reliability of this case.

\section{Discussion}
\label{sec:discusssion}

\begin{table}
    \centering
    \resizebox{\columnwidth}{!}{
    \begin{tabular}{l|c c c}
    \toprule
        \multirow{2}{*}{\textbf{Model}} & \multirow{2}{*}{\textbf{Eval. mode}} & \textbf{Video Motion} & \textbf{Physical Laws} \\
        ~ & ~ & \textbf{Acc / Pre / Cov / Con} & \textbf{Acc / Pre / Cov / Con} \\
        \midrule
        \multirow{2}{*}{GPT-4o} & auto & 11.3 / 47.7 / 88.8 / ~3.9~ & 26.0 / 63.4 / 96.3 / ~9.1~ \\
        ~ & human & 13.4 / 55.1 / 90.3 / ~4.6~ & 28.3 / 70.3 / 95.0 / ~9.9~ \\
        \midrule
        \multirow{2}{*}{Gemini 1.5 Pro} & auto & 10.4 / 41.7 / 91.5 / ~5.6~ & 27.9 / 62.8 / 97.2 / 15.0 \\
        ~ & human & 12.1 / 48.4 / 92.7 / ~6.5~ & 28.7 / 68.5 / 96.2 / 15.4 \\
        \midrule
        \multirow{2}{*}{Qwen2-VL-72B} & auto & ~7.3~ / 31.7 / 84.8 / ~4.7~ & 26.9 / 62.9 / 92.1 / 17.2 \\
        ~ & human & ~9.7~ / 42.1 / 86.2 / ~6.2~ & 28.3 / 68.5 / 94.3 / 18.1 \\
        \midrule
        \multirow{2}{*}{CogVLM2-Caption} & auto & ~6.0~ / 34.2 / 85.8 / ~3.9~ & 25.7 / 62.7 / 94.9 / 16.5 \\
        ~ & human & ~6.8~ / 39.7 / 87.2 / ~4.4~ & 26.8 / 65.5 / 96.2 / 17.2 \\
        \midrule
        \multirow{2}{*}{Tarsier-34B} & auto & ~7.7~ / 45.8 / 91.9 / ~8.6~ & 25.4 / 65.1 / 95.1 / 28.4 \\ 
        ~ & human & ~8.7~ / 51.3 / 90.2 / ~9.7~ & 27.8 / 69.9 / 93.7 / 31.1 \\
        \bottomrule
    \end{tabular}
    }
    \caption{Comparison between automated evaluation and human evaluation on VidCapBench-HE.}
    \label{tab:VidCapBench Consis Compare}
\end{table}

\noindent\textbf{Influence of the judge model}.~\label{sec:discuss of judge}
While GPT-4o provides stable judgments as a judge model, its cost and evaluation time pose challenges for iterative T2V development.
To investigate the impact of the judge model, we substitute GPT-4o with Qwen2-72B and Llama-3.3-70B as the judge model, using the same evaluation procedure with greedy decoding.
The results, summarized in Table~\ref{tab:VidCapBench Diff Judge Compare}, reveal significant discrepancies in evaluation outcomes. Notably, when Qwen2-72B is employed, we observe an average increase of 40 points in Pre score accompanied by a 35-point decrease in Cov score.
Such substantial discrepancies appear implausible and are inconsistent with human evaluations shown in Table~\ref{tab:VidCapBench human eval compare}.
Therefore, we caution against the use of locally deployed offline models for the judging process.
In contrast, Llama-3.3-70B demonstrates better alignment with GPT-4o, showing comparable Acc, Cov, and Con scores. However, it exhibits systematically higher Pre score that conflict with human judgments in Table~\ref{tab:VidCapBench human eval compare}.
Based on these findings, we recommend Llama-3.3-70B as a cost-effective alternative to GPT-4o when budget constraints are a concern, as it provides relatively reliable and meaningful evaluation results. Nevertheless, for more accurate and robust guidance during the development of new models, we still suggest utilizing stronger judge models like GPT-4o whenever feasible.

\noindent\textbf{Discrepancies between automated and human evaluation on VidCapBench-HE}.
In Section~\ref{sec:fine-grained analysis}, we have revealed the strong consistency between human and automatic evaluations on VidCapBench-AE, and meanwhile highlighting the significant inconsistencies on VidCapBench-HE. Here, we further elaborate on these discrepancies. As illustrated in Table~\ref{tab:VidCapBench Consis Compare}, notable discrepancies between automated and human evaluation on VidCapBench-HE are evident, with average differences of 1.6 in Acc, 6.1 in Pre, and 1.4 in Cov. Furthermore, the relative ranking of some models changes, which underscores the unreliability of automated evaluation under certain conditions. These results suggest that while automated evaluation may provide rapid feedback, human intervention is essential in more complex or nuanced scenarios to achieve a comprehensive and accurate assessment.

\begin{figure}
  \includegraphics[width=\linewidth]{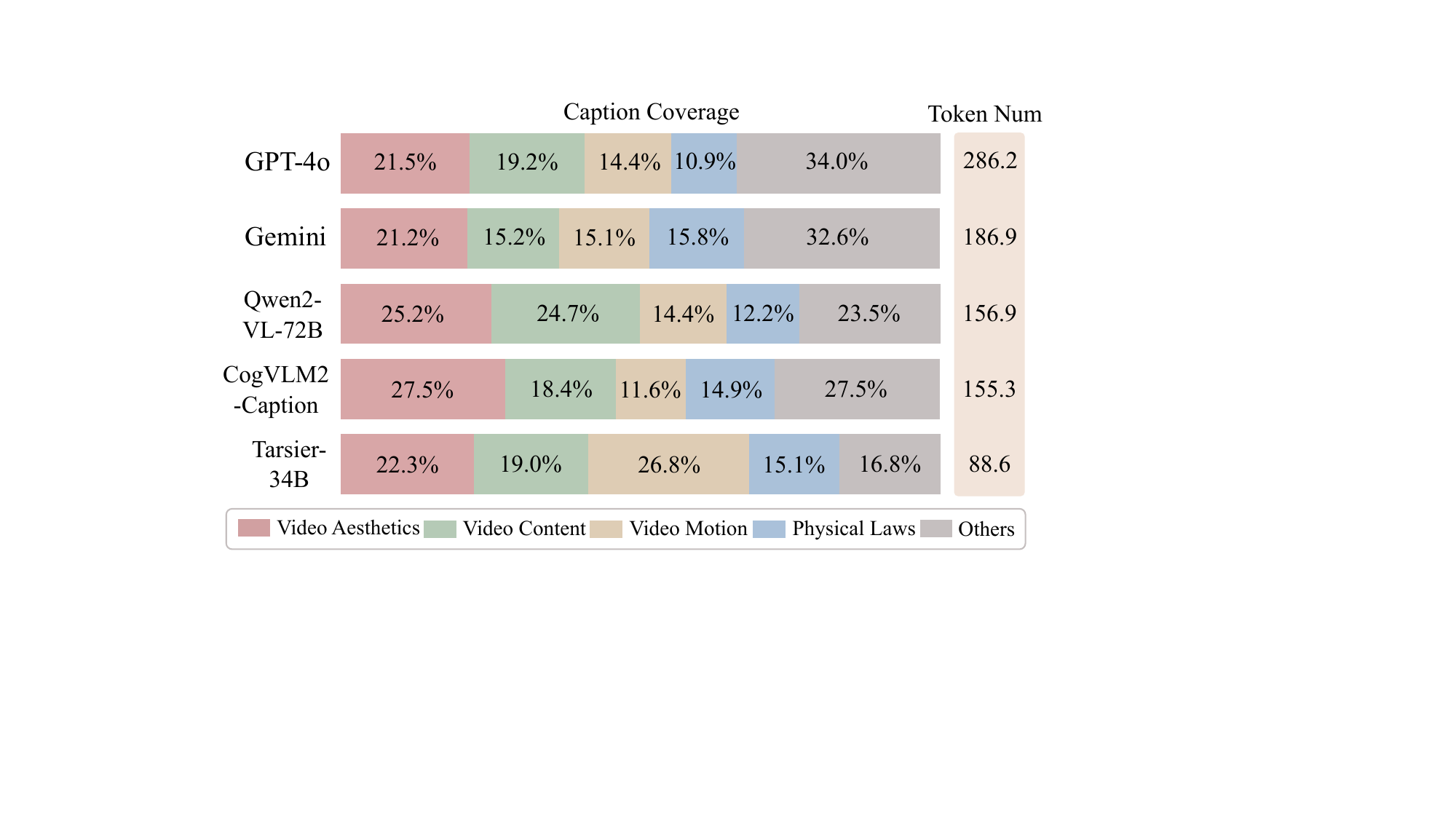}
  \caption{Distribution of caption coverage.}
  \label{fig:caption coverage}
\end{figure}
% \begin{table}
%     \centering
%     \resizebox{\columnwidth}{!}{
%     \begin{tabular}{l|cccccc}
%     \toprule
%         \textbf{Model} & \textbf{VA} & \textbf{VC} & \textbf{VM} & \textbf{PL} & \textbf{Others} & \textbf{T5 token num} \\
%         \midrule
%         GPT-4o-20240806 & 21.5\% & 19.2\% & 14.4\% & 10.9\% & 34.0\% & 286.2 \\
%         Gemini-1.5-Pro-002 & 21.2\% & 15.2\% & 15.1\% & 15.8\% & 32.6\% & 186.9 \\
%         Qwen2-VL-72B & 25.2\% & 24.7\% & 14.4\% & 12.2\% & 23.5\%  & 156.9 \\
%         CogVLM2-Caption & 27.5\% & 18.4\% & 11.6\% & 14.9\% & 27.5\% & 155.3 \\
%         Tarsier-34B & 22.3\% & 19.0\% & 26.8\% & 15.1\% & 16.8\% & 88.6 \\
%         \bottomrule
%     \end{tabular}
%     }
%     \caption {Distribution of caption coverage.}
%     \label{tab:caption coverage}
% \end{table}

\begin{figure*}[ht]
  \includegraphics[width=\linewidth]{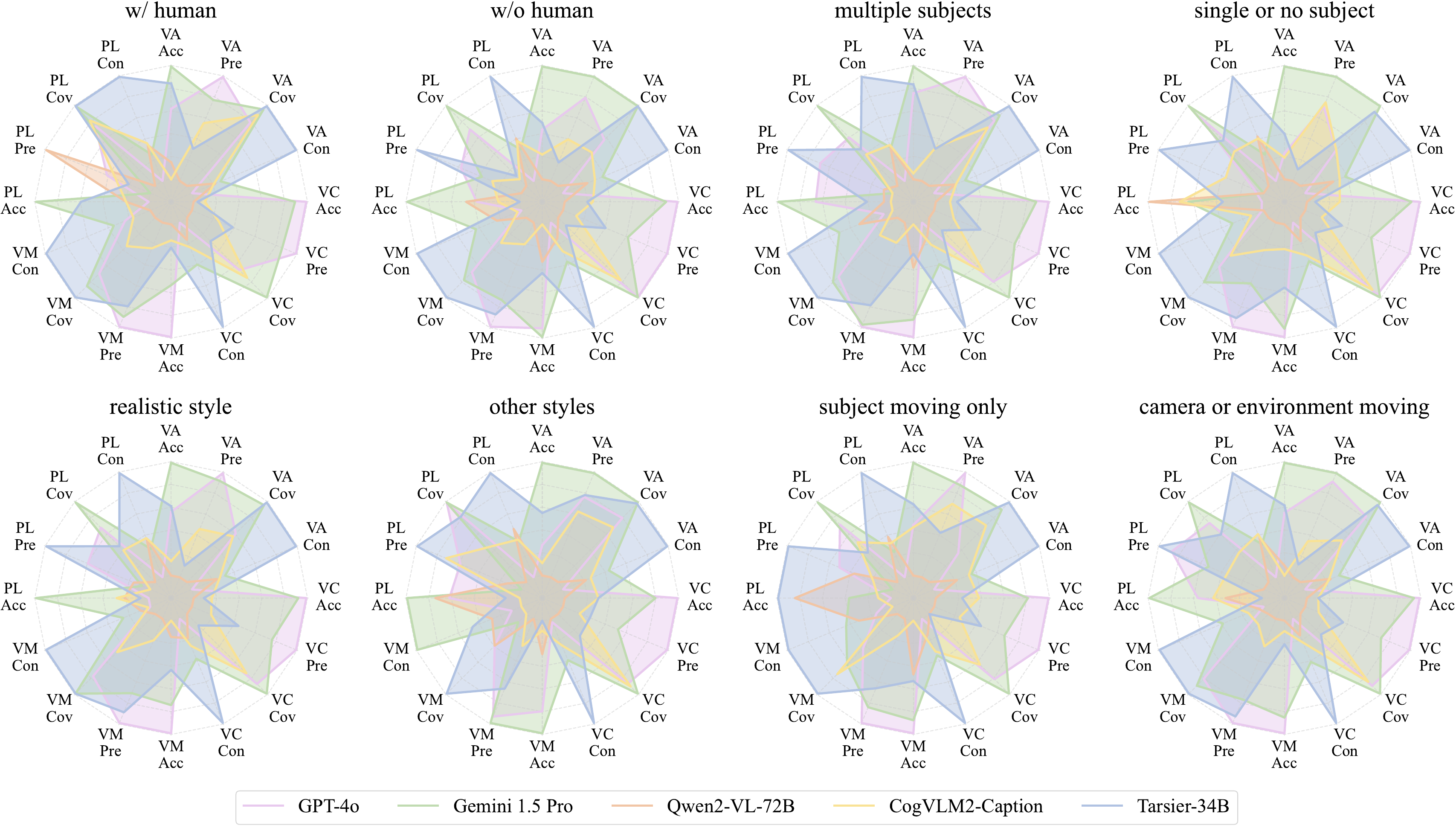}
  \caption{Performance comparison across diverse video categories in VidCapBench. ``VA'', ``VC'', ``VM'', and ``PL'' represent four evaluation dimensions, which are ``Video Aesthetics'', ``Video Content'', ``Video Motion'', and ``Physical Laws'', respectively. The absolute values of each dimension are normalized to facilitate clear comparison and visualization of relative performance differences.}
  \label{fig:rader}
\end{figure*}

\noindent\textbf{Caption coverage distribution}.
We analyze the coverage of four key categories within the generated captions, as depicted in Figure~\ref{fig:caption coverage}.
The models exhibit significant variations in caption coverage distribution.
GPT-4o produces the longest captions on average but also includes the highest proportion of irrelevant content, which may adversely affect T2V semantic responsiveness.
Tarsier-34B tends to output only a few events, resulting in shorter captions.
These observed differences in caption category coverage among the models are aligned with their respective performance on the focused dimensions within VidCapBench.
% These models differ in their coverage across categories, aligning with their respective performance on the focused dimensions in VidCapBench.

\noindent\textbf{Category-based evaluation}. To gain deeper insights into the capabilities of different models, we conduct a comprehensive analysis across distinct video categories within the full evaluation set. The categorization is based on four critical dimensions: (1) the presence of human figures, (2) the number of subjects, (3) visual styles, and (4) motion types. As presented in Figure~\ref{fig:rader}, our experimental results reveal several critical observations. Among all video categories, Gemini demonstrates superior performance in the dimension of Video Aesthetics, GPT-4o excels in the dimension of Video Content, and Tarsier outperforms others in the dimension of Video Motion and Physical Laws. Specifically, Tarsier performs exceptionally well when processing videos containing human figures, while its performance slightly declines when handling videos without human figures, where Gemini presents an overall good performance. In terms of the number of subjects, GPT-4o, Gemini, and Tarsier maintain robust performance regardless of the number of subjects in the scenes, whereas Qwen2-VL and CogVLM2 exhibit a noticeable decline when processing scenes with multiple subjects. Regarding video styles, Gemini continues to lead across different styles, while Qwen2-VL shows a performance drop when dealing with realistic styles. Notably, all evaluated models, except for GPT-4o, experience a decline in performance when handling challenging scenes involving camera or environmental movement.
These empirical findings highlight the varying strengths and limitations in current captioning models across different video categories, providing valuable insights for further improvements.

\end{document}